%% file: main.tex
\documentclass{article}

\usepackage{graphicx}
\usepackage{grffile}
\usepackage[nonatbib,preprint]{neurips_2024}

\usepackage[utf8]{inputenc} % allow utf-8 input
\usepackage[T1]{fontenc}    % use 8-bit T1 fonts
\usepackage{url}            % simple URL typesetting
\usepackage{booktabs}       % professional-quality tables
\usepackage{amsfonts}       % blackboard math symbols
\usepackage{nicefrac}       % compact symbols for 1/2, etc.
\usepackage{microtype}      % microtypography
\usepackage{xcolor}         % colors
\usepackage{tcolorbox}
\usepackage{times}
\usepackage{epsfig}
\usepackage{amsmath}

\usepackage{array}
\usepackage{subcaption}
\usepackage{multirow,multicol}
\usepackage[flushleft]{threeparttable}
\usepackage{comment}
\usepackage{algorithmic}
\usepackage{booktabs}
\usepackage{multicol}
\usepackage{multirow}
\usepackage{bbm, dsfont}
\usepackage[font=small,labelfont=bf]{caption}

\usepackage{amssymb}% http://ctan.org/pkg/amssymb
\usepackage{pifont}% http://ctan.org/pkg/pifont

\newcommand{\app}{\raise.17ex\hbox{$\scriptstyle\sim$}}

\newlength\savewidth

\makeatletter\renewcommand\paragraph{\@startsection{paragraph}{4}{\z@}
  {.5em \@plus1ex \@minus.2ex}{-.5em}{\normalfont\normalsize\bfseries}}\makeatother

\def\tablecite#1#{%
  \def\pretablecite{#1}%
  \tableciteaux}
\def\tableciteaux#1{%
  \textsuperscript{\expandafter\originalcite\pretablecite{#1}}%
}

\usepackage{graphicx}
\usepackage{enumitem}
\usepackage{wrapfig}
\usepackage{lipsum}

\newcolumntype{H}{>{\setbox0=\hbox\bgroup}c<{\egroup}@{}}
\newcolumntype{a}{>{\columncolor{Gray}}c}

\usepackage{tabu}
\usepackage{nicematrix}

\definecolor{ForestGreen}{rgb}{0.13, 0.55, 0.13}
\definecolor{Green}{rgb}{0.0, 0.5, 0.0}
\definecolor{green(munsell)}{rgb}{0.0, 0.66, 0.47}
\definecolor{green(ryb)}{rgb}{0.4, 0.69, 0.2}
\definecolor{green(pigment)}{rgb}{0.0, 0.65, 0.31}
\definecolor{citecolor}{HTML}{0071bc}
\definecolor{GrayXMark}{gray}{0.7}

\usepackage{tabularx}
\usepackage[export]{adjustbox}

\definecolor{ForestGreen}{rgb}{0.13, 0.55, 0.13}
\definecolor{Green}{rgb}{0.0, 0.5, 0.0}
\definecolor{green(munsell)}{rgb}{0.0, 0.66, 0.47}
\definecolor{green(ryb)}{rgb}{0.4, 0.69, 0.2}
\definecolor{green(pigment)}{rgb}{0.0, 0.65, 0.31}

\newcolumntype{x}[1]{>{\centering\let\newline\\\arraybackslash\hspace{0pt}}p{#1}}
\definecolor{Gray}{gray}{0.9}

\usepackage{makecell}

\definecolor{ForestGreen}{rgb}{0.13, 0.55, 0.13}
\definecolor{Green}{rgb}{0.0, 0.5, 0.0}
\definecolor{green(munsell)}{rgb}{0.0, 0.66, 0.47}
\definecolor{green(ryb)}{rgb}{0.4, 0.69, 0.2}
\definecolor{green(pigment)}{rgb}{0.0, 0.65, 0.31}
\definecolor{citecolor}{HTML}{0071bc}
\definecolor{GrayXMark}{gray}{0.7}

\usepackage{tabularx}
\usepackage[export]{adjustbox}

\usepackage{hyperref}
\hypersetup{
    colorlinks,
    linkcolor={black},
    citecolor={black},
    urlcolor={black}
}
\usepackage[capitalize]{cleveref}
\crefname{section}{Sec.}{Secs.}
\Crefname{section}{Section}{Sections}
\Crefname{table}{Table}{Tables}
\crefname{table}{Table}{Tabs.}

\definecolor{ForestGreen}{rgb}{0.13, 0.55, 0.13}
\definecolor{Green}{rgb}{0.0, 0.5, 0.0}
\definecolor{green(munsell)}{rgb}{0.0, 0.66, 0.47}
\definecolor{green(ryb)}{rgb}{0.4, 0.69, 0.2}
\definecolor{green(pigment)}{rgb}{0.0, 0.65, 0.31}

\usepackage{colortbl}
\usepackage{xspace}
\newcommand{\ours}{PopAlign\xspace}

\usepackage{color, colortbl}

\usepackage{graphicx}
\usepackage{wrapfig}

\title{PopAlign: Population-Level Alignment for \\Fair Text-to-Image Generation }

\author{Shufan Li, Harkanwar Singh, Aditya Grover \\
\{jacklishufan,harkanwarsingh,adityag\}@cs.ucla.edu \\
University of California, Los Angeles
}

\newcommand{\name}{PopAlign}

\begin{document}
\maketitle

\begin{abstract}
Text-to-image (T2I) models achieve high-fidelity generation through extensive training on large datasets. However, these models may unintentionally pick up undesirable biases of their training data, such as over-representation of particular identities in gender or ethnicity neutral prompts. Existing alignment methods such as Reinforcement Learning from Human Feedback (RLHF) and Direct Preference Optimization (DPO) fail to address this problem effectively because they operate on pairwise preferences consisting of individual \textit{samples}, while the aforementioned biases can only be measured at a \textit{population} level. For example, a single sample for the prompt ``doctor" could be male or female, but a model generating predominantly male doctors even with repeated sampling reflects a gender bias. To address this limitation, we introduce PopAlign, a novel approach for population-level preference optimization, while standard optimization would prefer entire sets of samples over others. We further derive a stochastic lower bound that directly optimizes for individual samples from preferred populations over others for scalable training.
% We derive this alignment objective by directly optimizing for preferred population samples over unpreferred ones. 
% that extends RLHF objective by incorporating a reward model based on population-level traits. We show that this extended population-level RL objective can be converted to a supervised objective on individual samples, allowing for efficient training. 
Using human evaluation and standard image quality and bias metrics, we show that PopAlign significantly mitigates the bias of pretrained T2I models while largely preserving the generation quality. Code is available at \hyperlink{https://github.com/jacklishufan/PopAlignSDXL}{https://github.com/jacklishufan/PopAlignSDXL}.
% T2i Models made great progess, from large scale pretraing data
% T2i models may have bias. E.g. white, male
% RLHF/DPO used align generative models, by contrasting pair of preferred and unpreferred data 
% existing alignment methods fail to address this bias because they only operate on sample level. For example, we cannot say "female doctor is more diverse", as "all femable doctor" is not diversity.
% To bridge this gap, we propose PopAlign which align diffusion model based on population level traits. PopAlign extends the standard RLHF formulation and introduce pop-level reward. Using similar techniques as DPO, we convert it to supervised objective for efficient trianing
% From human evaluation and automatic metric, results show that PopAlign signifcantly mitigate the bias of pretrained T2I model with minimal degradion in generation quality. 
% Limitations: while PopAlign can by no means eliminate all biases, it offers a effective way to mit

\end{abstract}

\input{sections/1_introduction}

\input{sections/2_related_works}
\input{sections/3_method}
\input{sections/4_results}

\input{sections/5_conclusion}

%%%%%%%%%%%%%%%%%%%%%%%%%%%%%%%%%%%%%%%%%%%%%%%%%%%%%%%%%%%%

% \clearpage
% \resetlinenumber
% \setcounter{page}{1}

\newpage
\clearpage

{\small
\bibliographystyle{abbrv}
\bibliography{references}
}

\newpage
\clearpage

\input{sections/6_appendix}
\end{document}

%% file: sections/1_introduction.tex
% \def\figPipeline#1{
%     \captionsetup[sub]{font=small}
%     \begin{figure*}[#1]
%       \centering
%       \includegraphics[width=1.0\linewidth]{figures/pipeline.pdf}
%       \vspace{-3pt}
%       \caption{
%       Network architecture of SegLLM for interactive image reasoning segmentation. 
%       }
%       \label{fig:teaser}
%     \end{figure*}
% }

%       }
%       \label{fig:teaser}
%     \end{figure*}
% }

\section{Introduction} 
\begin{figure}[t]
    \centering
    \begin{subfigure}[b]{0.48\textwidth}
    \centering
 \includegraphics[width=1.0\linewidth]{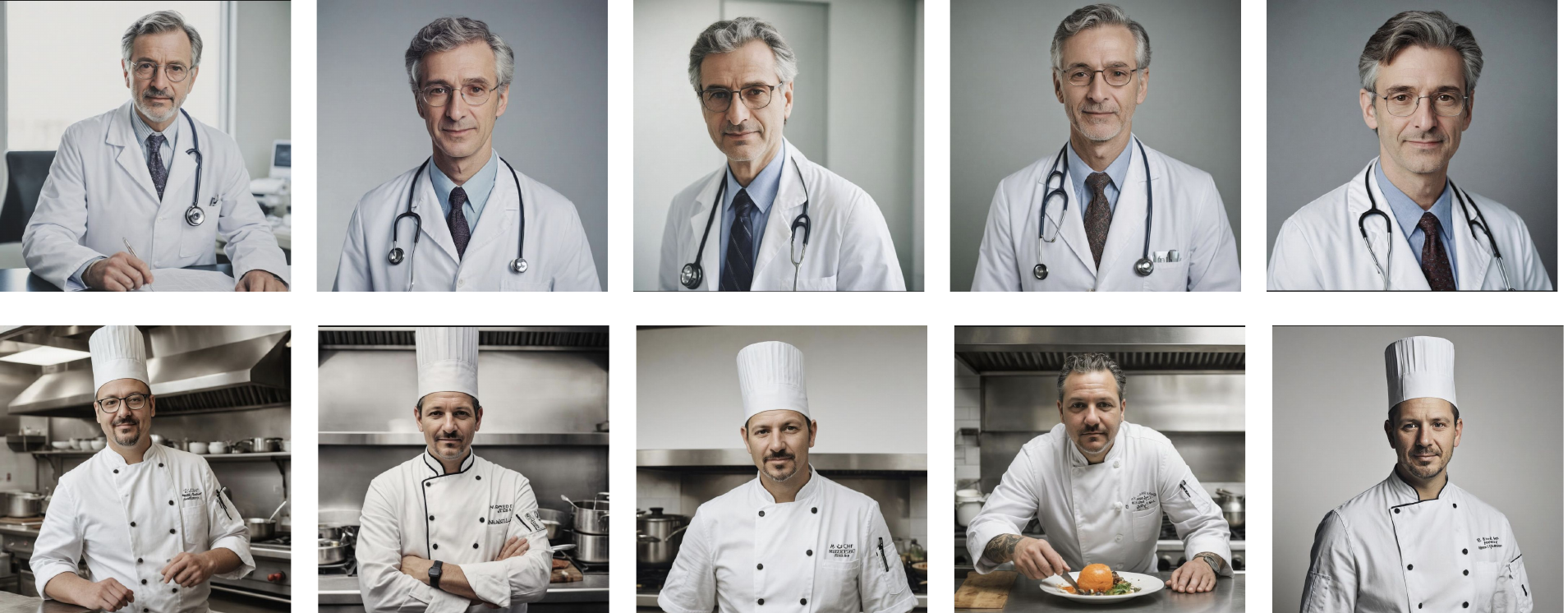}
    \caption{SDXL }
       \label{fig:teaser-1a}
    \end{subfigure}
    ~~~
    \begin{subfigure}[b]{0.48\textwidth}
    \centering
        \includegraphics[width=1.0\linewidth]{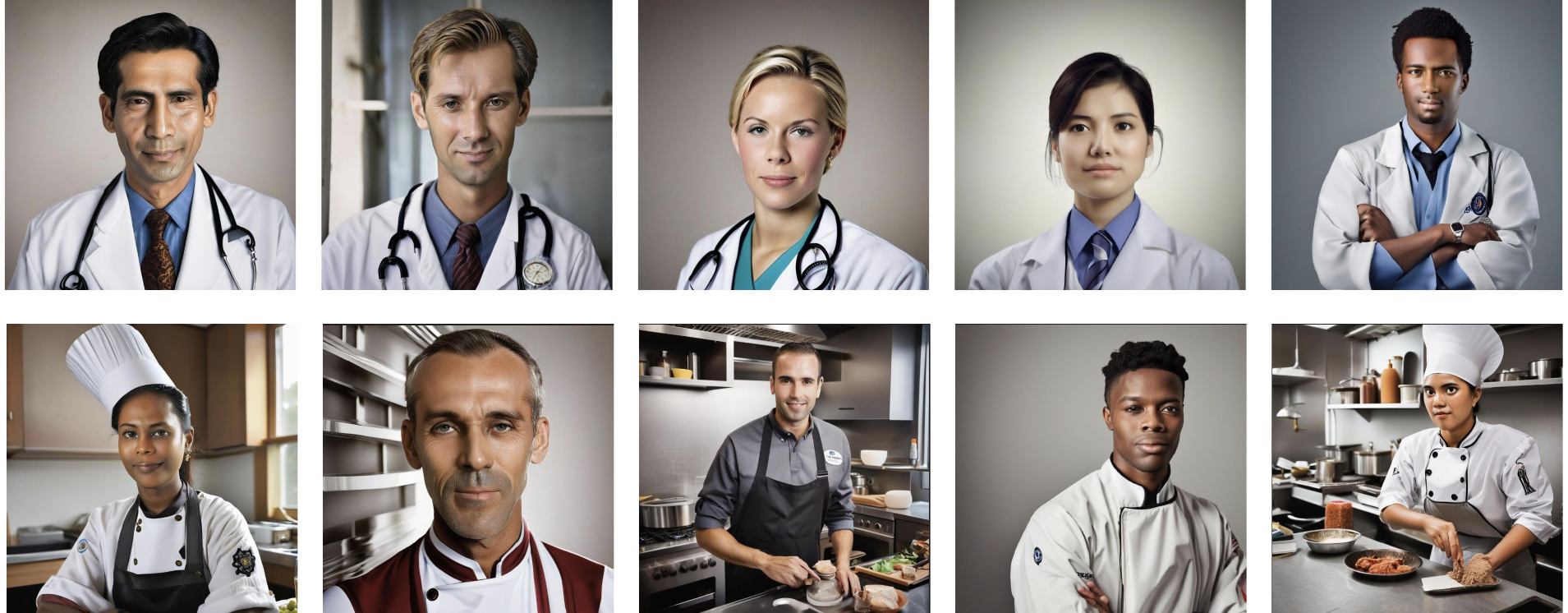}
    \caption{\ours~ }
    \end{subfigure}
   
\caption{Illustration of \ours, our proposed framework for mitigating the bias of pretrained T2I models using population-level alignment. \textbf{Left:} SDXL over-represents a particular identity as it picked up biases of the training data.  \textbf{Right:} \ours mitigates the biases without compromising the quality of generated samples. }

\end{figure}

Modern image generative models, such as the Stable Diffusion \cite{rombach2022high,stabilityai} and DALLE~\cite{ramesh2021zero,ramesh2022hierarchical,openai2023dalle2} model series, are trained on large datasets of billions of images scraped from the Internet.
As a result, these models tend to strongly inherit various kinds of biases in their dataset. For example, in Figure \ref{fig:teaser-1a}, we can see that SDXL  tends to generate predominantly male images for the prompt "doctor," amplifying underlying societal biases as these models make their ways into an increasing number of everyday products and applications.
Several past works have documented such societal biases for foundation models at large \cite{bias1,bias2}, yet mitigation efforts lag, especially for text-to-image generation.

% female images for prompts such as "teacher" and "cashier," and 

% Notably, one key feature of the aforementioned biases is that they can only be assessed 
In this work, we study a specific category of biases that are defined at a \textit{population} level. 
That is, a single sample from a generative model is insufficient to assess whether the model exhibits a specific population bias.
Prominent examples include biases of text-to-image generative models with respect to gender or ethnicity neutral prompts.
For example, a single generated image sample for the prompt ``doctor" could be male or female, but a model generating images of predominantly male doctors even with repeated sampling reflects a gender bias.
This is in contrast with much of the AI safety and alignment work in recent times for large language models \cite{dai2023safe,zhang2024shieldlm}, where the harmfulness in generations can be ascertained at the level of individual samples. For example, given the prompt "what is the gender of doctors?", even individual generated text responses should ideally not show a bias towards a specific gender.

Given any implicit population preference (e.g., equalizing image generations across genders for a gender-neutral prompt), there are two key challenges in aligning large-scale text-to-image generative models. 
First, many state-of-the-art models are trained on large-scale, possibly non-public datasets, making it prohibitively expensive for intermediate developers to retrain them for population alignment. Therefore, an ideal solution would build on existing models, be sample-efficient in acquiring additional supervision, and parameter-efficient for cost-effective alignment.
Second, given the diverse range of concepts represented in modern generative models, population alignment on a specific dimension (e.g., gender) should not degrade visual quality for any kind of prompt.
Given these criteria, we also note that prior works \cite{choi2020fair,tan2020improving,teo2023fair,um2023fair} involving retraining small-scale generative models trained on narrow datasets (e.g.,  CelebA) with data re-sampling or class-balancing loss cannot be directly applied because in our setting, the pretraining data can be very large or unavailable, and visual quality is evaluated more broadly over a wider range of prompts. 

Our primary contribution in this work is to define \name{}, a preference alignment framework for mitigating population bias for text-to-image generative models. Standard preference alignment frameworks, such as reinforcement learning from human preferences (RLHF)~\cite{christiano2017deep,ouyang2022training} and its reward-free extension direct preference optimization (DPO)~\cite{rafailov2024direct}, cannot be directly applied for mitigating population bias as they acquire either absolute ratings or pairwise preferences between individual samples.
For image generation, this information is only helpful for improving visual quality or semantic adherence to prompts, as shown in recent works \cite{wallace2023diffusion}.
Building on the reinforcement learning from human preferences (RLHF) framework, we first propose to acquire multi-sample preferences over sets of samples, as proxies for population-level preferences. 
We reduce it to a corresponding reward-free, population-level DPO objective.
Finally, we derive the \name{} objective as a stochastic lower bound to this population-level DPO objective such that it permits tractable evaluation and maximization by decomposing multi-sample pairwise preferences into single-sample preferences after sampling from their respective populations.
% the preferred and unpreferred populations.

 \begin{figure}[t]
    \centering
    \begin{subfigure}[b]{0.48\textwidth}
    \centering
 \includegraphics[width=1.0\linewidth]{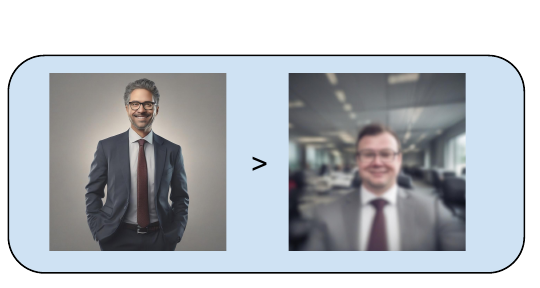}
    \caption{Sample-level preferences (RLHF/DPO)}
    \end{subfigure}
    \begin{subfigure}[b]{0.48\textwidth}
    \centering
        \includegraphics[width=1.0\linewidth]{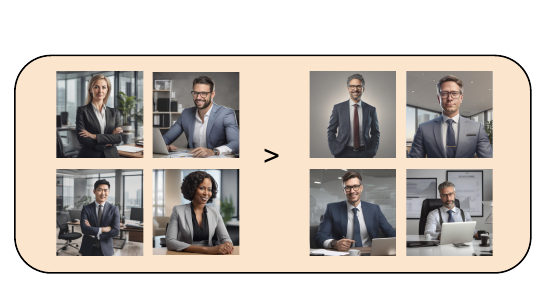}
    \caption{Population-level preferences (\ours)}
    \end{subfigure}
     \label{fig:explainer}
\caption{Difference between PopAlign and existing RLHF/DPO Methods. \textbf{Left:} Existing methods such as RLHF/DPO use pairwise preferences of individual samples to improve image quality. \textbf{Right} \ours uses population-level preferences to achieve better fairness and diversity.}
\end{figure}

To evaluate our model's efficacy, we collect population-level preference data through a combination of human labelers and automatic pipelines based on attribute classifiers. Through standard image quality and bias metrics as well as extensive human evaluations, we show that \ours significantly mitigates bias in pretrained text-to-image models without notably impacting the quality of generation. Compared with a base SDXL model, \ours reduces the gender and race discrepancy metric
of the pretrained SDXL by (-0.233), and (-0.408) respectively, while maintaining comparable image quality.

% In summary, we proposed \

\label{sec:intro}

%% file: sections/2_related_works.tex
\section{Related Works}

\subsection{Diversity and fairness in image generation}

Diversity and fairness are active areas of research in image generation.
However, these terminologies often refer to distinct concepts in past works. The word diversity is used to refer broadly to the coverage of concepts in the training distribution. Accordingly, many techniques exist to improve diversity. For example, in current diffusion models, we can tune the  guidance \cite{dhariwal2021diffusion,ho2022classifier} as a knob for trading off diversity with image quality. However, these works as well as recent extensions (e.g., \cite{kim2022refining},  \cite{sadat2023cads}) focus on diversity as a generic term, and not diversity of specific attributes that have fairness and equity implications such as race and gender. For examples, for the prompt "doctor", a set of images of white male doctors with varying hairstyles, camera angles, lighting conditions, backgrounds can be considered as more "diverse" than generating a single image of a middle-aged doctor with the same pose and background. While indeed diverse along one axis, this notion does not capture "fair" representation of identities, which is the focus of this work.

Another line of related works focus on "fairness", which measures whether generative samples matches a desired distribution over a specific sets sensitive attributes such as gender and race. We discuss some representative works Early  approaches that reweigh the importance of samples in a biased training dataset to improve fairness \cite{choi2020fair}. FairGen \cite{tan2020improving} improve the fairness of a pretrained Generative adversarial network (GAN) by shifting its latent distribution using Gaussian mixure models. FairTL~\cite{teo2023fair} improves the fairness of GAN by fine-tuning a discriminator on a small unbiased dataset. Um and Suh \cite{um2023fair} employs LC-divergence to improve the fairness of GAN, which better captures  the distance between real and generated in small training datasets. Despite their successes, these methods are tested on small datasets such as CelebA \cite{liu2015deep}. They are not applicable to T2I diffusion models pretrained on large-scale datasets either because of GAN specific designs or requires re-training using the pretrained data. Most recently, FairDiffsuion attempts to mitigate the bias of diffusion model at \textit{deployment} time by maintaining a lookup table of known biases (e.g. "prompt doctor is biased towards male"), and injecting image-editing prompts at inference time. By contrast, our method address the problem at model \textit{release}. It is also more flexible and does require maintaining a lookup table of known biases.

\subsection{Aligning generative models with human preferences} 

A growing line of recent work considers the \textit{alignment} of the outputs of large language models (LLMs) to improve their safety and helpfulness by directly querying humans (or other AI models) to rank or rate model outputs to create a preference dataset.
% The problem of alignment initiallize arise with fine-tuning large-language models (LLMs), where the goal is to improve the safety and helpfulness of a pretrained LLM by training on a small preference dataset without sacrificing the capabilities gained through large-scale pretraining. 
The most basic approach is reinforcement learning with human preferences (RLHF) \cite{christiano2017deep}, which trains a reward model on this preference data and then employs reinforcement learning to maximize the expected rewards. The RL step typically make use of proximal policy optimization (PPO) \cite{schulman2017proximal} to prevent the model from diverging too much from the pretrained model. DPO \cite{rafailov2024direct} simplified this process by converting the RL objective to a supervised-finetuning-style objective, eliminating the need to first fit a reword model. Recently, various works \cite{wallace2023diffusion,yang2023using}  extended DPO to text-to-image diffusion models. These works mostly focus on improving the quality of generated images, with little emphasis on fairness and safety.

% For example, the Pick-a-Pick dataset used by Diffusion-DPO is known to have NSFW prompts and images. 

% is used
% TLDR: 1. RLHF,DPO used in LLM. 2. Diffusion DPO applied it to diffsion models. 
% 3. In general, aligning diffusion model is under explored. 

\label{sec:rworks}

%% file: sections/3_method.tex
\section{Background}
 \subsection{Reinforcement Learning with Human Feedback}
\label{sec:background-rlhf}
Reinforcement Learning with Human feedback (RLHF) aims to fine-tune a generative model to better align its outputs with human preferences. Following the collection of a dataset of human preferences, RLHF is typically a two-stage training process. In the first stage, a reward model is fitted on collected pairwise preferences using the Bradley-Terry (BT) model \cite{bradley1952rank} with the following cost function:
\begin{equation}
    \mathcal{L}_r(r,\mathcal{D}))=-\mathbb{E}_{c, x^w, x^l \sim \mathcal{D}} [\text{log} \sigma ( r(x_w,c) - r(x_l,c))]
    \label{eq:reword_fitting}
\end{equation}
where $r$ is a reward model parameterized by a neural network, $x_w$ and $x_l$ are "winning" and "losing" samples in a pair, and $c$ is the input prompt. In the second stage, the generative model $\pi_\theta$ is fine-tuned to maximize the expected reward of generated samples using Proximal Policy Optimization (PPO), while ensuring  $\pi_\theta$ do not change too much from the initialization point $\pi_\text{ref}$. This is achieved by incorporating a KL-term as a divergence penalty in the following objective:
\begin{equation}
    \underset{\pi_{\theta}}{\text{max }} \mathbb{E}_{c \sim \mathcal{D}, x\sim \pi_{\theta}(x|c)} [r(x,c)] -\beta \mathbb{D}_{\text{KL}}[\pi_{\theta}(x|c) ||\pi_{\text{ref}}(x|c) ]
    \label{eq:max_reward}
\end{equation}
where $\beta$ is a hyper-parameter controlling the strength of divergence penalty.

\subsection{Direct Preference Optimization}
\label{sec:background-dpo}

% showed that the optimal solution of \cref{eq:max_reward} $\pi_\theta^*$ satisfy the following equation

% \begin{equation}
% r^*(x,c) = \beta \log \frac{\pi_\theta^*(x|c)}{\pi_{\text{ref}}(x|c)} + \beta \log Z(c)
% \end{equation}

% where $Z(c)$ is the partition function. Substituting $r^*(x,c)$ in \cref{eq:max_reward2}, we can obtain an equivalent objective 

One major challenge in training the model with RLHF is that the two stage training process can be unstable and computationally inefficient.  Direct Preference  Optimization (DPO) address this by \cite{rafailov2024direct} converting the RLHF formulation to a supervised objective without an explicit reward model:
\begin{equation}
    \underset{\pi_{\theta}}{\text{max }}\mathbb{E}_{x^w,x^l,c \sim \mathcal{D}}[\log \sigma(\beta \log \frac{\pi_\theta(x^w|c)}{\pi_{\text{ref}}(x^w|c)}-\beta \log \frac{\pi_\theta(x^l|c)}{\pi_{\text{ref}}(x^l|c)})]
    \label{eq:loss_dpo_vanilla}
\end{equation}
In this setup, an implicit reward model $r(x,c) = \beta \log \frac{\pi_(x|c)}{\pi_{\text{ref}}(x|c)} + \beta \log Z(c)$ is used, where $Z(c)$ is the partition function. 

\subsection{Diffusion models}
\label{sec: background-diffusion}
Denoising Diffusion Probabilistic Models (DDPM) \cite{ho2020denoising} use a Markov chain to model the image generation process starting from i.i.d white noises. The forward diffusion process $p(x_{t+1}|x_{t})$ gradually adds noise to an image $x_t$ at timestamp $t$ according to a noise schedule, until it converts the initial noise-free image $x_0$ to i.i.d. Gaussian noise $x_T$. A generative diffusion model can be trained to fit the reverse process $q_\theta(x_{t-1}|x_t)$ using the evidence lower bound (ELBO) objective:
 \begin{equation}
    \mathcal{L_{\text{DDPM}}}=\mathbb{E}_{x_0, t, \epsilon}[\lambda(t)\lVert\epsilon_t-\epsilon_\theta(x_t,t)\rVert^2]
\end{equation}
where $\lambda(t)$ is a time-dependent weighting function dependent on the noise schedule, $\epsilon_t$ is the added noise at time stamp $t$, and $\epsilon_\theta$ is the diffusion model parameterized by $\theta$. In the sampling process, we start at i.i.d Gaussian noise $x_T$ and gradually remove the noise, until reaching the final image $x_0$.

\subsection{Diffusion-DPO}
The DPO framework can also be extended to diffusion models. A key challenge in applying the DPO objective in \cref{eq:loss_dpo_vanilla} to diffusion models is that the conditional probability $\pi(x_0|c)$ can only be computed by marginalizing over all possible sampling trajectories $x_{0:T}$, which is infeasible. Diffusion-DPO \cite{wallace2023diffusion} resolve this by defining a reward model dependent on a specific chain $x_{0:T}$, rather than depending on the final sample $x_0$ only. This leads to the following objective
\newcommand{\bbE}{\mathbb{E}}
\newcommand{\calD}{\mathcal{D}}
\newcommand{\xw}{x^w}
\newcommand{\xl}{x^l}
\newcommand{\pref}{p_\text{ref}}
\begin{equation}
    \max_{\pi_{\theta}}\bbE_{(\xw_{0},\xl_0)\sim\calD }
    \log \!\sigma\! \biggl( \beta
    {\mathbb{E}}_{{\substack{\xw_{1:T} \sim p_\theta(\xw_{1:T}|\xw_0) \xl_{1:T} \sim p_\theta(\xl_{1:T}|\xl_0)} }} 
    \left[ \log\! \frac{p_{\theta}(\xw_{0:T})}{\pref(\xw_{0:T})}\! -\! \log\! \frac{p_{\theta}(\xl_{0:T})}{\pref(\xl_{0:T})}\right]\biggr).
\end{equation}
Using Jensen’s inequality, Diffusion-DPO \cite{wallace2023diffusion} derived and optimized a tractable lower bound:
\begin{equation}
\max_{\pi_{\theta}}\bbE_{\substack{{(\xw_{0},\xl_0)\sim\calD}, t \sim \mathcal{U}(0,T)}} 
    \log\sigma \left(\beta T 
     \log\! \frac{p_{\theta}(\xw_{t-1}|\xw_{t})}{\pref(\xw_{t-1}|\xw_{t})}\! - \! \beta T \log\! \frac{p_{\theta}(\xl_{t-1}|\xl_{t})}{\pref(\xl_{t-1}|\xl_{t})}\right).\label{eq:loss-diffusion-dpo}
\end{equation}
Eq.~\ref{eq:loss-diffusion-dpo} allows efficient training without sampling through the whole reverse process for each update. 

\section{Method}
Consider a pretrained text-to-image model $\pi_\theta$ that is biased w.r.t. one or more population-level traits. Our goal in population-level alignment is to fine-tune \name{} \textit{without} acquiring any additional real images. To do so, we assume access to a source of preferences (e.g., via humans) over the model's output generations. 
% Since we are interested in population-level alignment, we propose to acquire preferences that compare two \textit{sets} of generated samples, as described next. 

\subsection{Population-Level Preference Acquisition}
\label{sec: data_pipeline}
Typically, alignment data for RLHF/DPO is created by generating multiple samples using the same prompt and asking humans to rank the results. Since the goal of \ours is to mitigate the population-level bias, we need to generate two or more \textit{sets} of images for the same prompt. However, naive sampling of sets does not work due to the high degree of bias within current T2I models for identity-neutral prompts.
% One of the major challenges in generating a good visual preference dataset for bias mitigation is that generating a balanced set is hard given the diffusion models are so biased. 
For example, we observe that among 100 images generated from the prompt "doctor", only 6 are female doctors. In the extreme case, when prompted with the prompt "engineer", the model generates no images of female engineers amongst 100 samples. This makes generating a set of near-fair samples nearly impossible using this naive method. 

To address this challenge, we use an approximated process where we directly augment a gender-neutral prompt such as "engineering" to a diverse set of identity-specific prompts such as "Asian male engineer" and "female engineer", and use images sampled from these augmented prompts as the \textit{winning set}, and images sampled directly from the gender-neutral prompt as the \textit{losing set}. As a sanity check, for each pair of sets, we use a classifier in combination with a face detector to determine if the sampled images are indeed consistent with the prompts. We drop pairs that are incorrect or ambiguous and fails this check. For example, we found that many images generated with the prompt "astronaut" contains a person with helmet on, making it impossible to determine the gender or ethnicity. These samples fail the detector and are dropped from the preference dataset.

% \ag{i think the above para is not consistent with the protocol below. above para is saying we augment gender neutral prompts with underrep prompts. para below is saying we "replace" overrep images with underrep images. even though eventually they many imply the same distribution. i think currently this para below might cause more confusion for reviewers. i would consider commenting it out}
% While this process may seem a clear departure from existing human-in-the-loop process of creating preference datasets, we argue that it is akin to running a Markov chain with a human-prompted preference kernel.
% approximation of Markov chain Monte Carlo (MCMC) sampling with human aids. 

% \todonote{We use MCMC, not Gibbs sampling to motivate because it may raise questions on if the samples are i.i.d. Gibbs sampling need conditional probability, however if the samples are i.i.d it doesnt really make sense?  }

\subsection{Population-Level Alignment from Human Preferences}

Given a prompt $c$ and two sets of generated images $X_0,X_1$ where $|X_0|= |X_1| = N$, The Bradley-Terry (BT) model \cite{bradley1952rank} for human preference is $p^*(X_0 \succ X_1 | c) = \sigma ( r(X_0,c) - r(X_1,c)) $,
where $r(X,c)$ is a real-valued reward function dependent on the prompt and the set of generated images. 

In the RLHF setup~\cite{ouyang2022training}, $r(X,c)$ is modeled by a neural network $\phi$ trained on a dataset $\mathcal{D}$ with pairs of winning samples and losing samples $(X^w, X^l, c)$ by optimizing the following objective function: 
\begin{equation}
    \mathcal{L}_r(r_\phi,\mathcal{D}))=-\mathbb{E}_{c, X^w, X^l \sim \mathcal{D}} [\text{log} \sigma ( r(X_w,c) - r(X_l,c))].
    \label{eq:loss_reward_model}
\end{equation}
Once the reward model is trained, we can optimize a generative model $\pi_\theta$ using the PPO objective:
\begin{equation}
    \underset{\pi_{\theta}}{\text{max }} \mathbb{E}_{c \sim \mathcal{D}, x_1,..x_N\sim \pi_{\theta}(x|c)} [r(\{x_1,..x_N\},c)] -\beta \mathbb{D}_{\text{KL}}[\pi_{\theta}(X|c) ||\pi_{\text{ref}}(X|c) ]
    \label{eq:rlhf_loss}
\end{equation}
where $X={x_1,..x_N}$ is a population of generated samples and $\pi_\text{ref}$ is a reference distribution. Typically, $\pi_\text{ref}$ is a pretrained model and $\pi_\theta$ is initialized with $\pi_\text{ref}$.
% \subsection{Direct Preference Optimization}
Further, using an analogous derivation as DPO \cite{rafailov2024direct}, we know that the optimal solution of \cref{eq:rlhf_loss}, say $\pi_\theta^*$ satisfies the condition $r^*(X,c) = \beta \log \frac{\pi_\theta^*(X|c)}{\pi_{\text{ref}}(X|c)} + \beta \log Z(c)$, where Z(c) is the partition function. Combining this with \cref{eq:loss_reward_model},  we obtain an equivalent objective:
\begin{equation}
    \underset{\pi_{\theta}}{\text{max }}\mathbb{E}_{c,X^w,X^l \sim \mathcal{D}}[\log \sigma(\beta \log \frac{\pi_\theta(X^w|c)}{\pi_{\text{ref}}(X^w|c)}-\beta \log \frac{\pi_\theta(X^l|c)}{\pi_{\text{ref}}(X^l|c)})].
    \label{eq:dpo_loss}
\end{equation}
Using this objective, we can directly optimize $\pi_\theta$ without explicitly training a reward model. 

\subsection{Population Level Alignment of Text-to-Image Diffusion Models}
\label{sec:main_derivation}

In the context of text-to-image diffusion models, the winning and losing population  $X^w,X^l$ each consists of $N$ images generated independently through the diffusion process $\{x^{w,i}\}_{i=1,2..N},\{x^{l,i}\}_{i=1,2..N}$. Hence, we can rewrite \cref{eq:dpo_loss} as:
\begin{equation}
    \underset{\pi_{\theta}}{\text{max }}\mathbb{E}_{c,X^w,X^l \sim \mathcal{D}}[\log \sigma(\beta \log \frac{\prod_{i=1}^N\pi_\theta(x^{w,i}|c)}{\prod_{i=1}^N\pi_{\text{ref}}(x^{w,i}|c)}-\beta \log \frac{\prod_{i=1}^N\pi_\theta(x^{l,i}|c)}{\prod_{i=1}^N\pi_{\text{ref}}(x^{l,i}|c)})].
        \label{eq:dpo_loss_pop}
\end{equation}
% \ag{this is the key mathematical novelty. just saying "using similar derivation as diffusion-dpo" is confusing (makes it sound that pop alignment is same as alignment notion in diffusion-dpo) and diminishes our contrib (we have an extra lower bound step over samples in X. very easy to miss that novelty the way the text is currently written). expand on this derivation with a few more key steps.}
 Naively using this objective can be computationally expensive, because it requires computing the distribution of all samples in the set at the same time. However, we can further establish a lower bound of this objective by applying Jensen’s inequality on the concave function $\log\sigma(x)$:
 \begin{equation}
    \underset{\pi_{\theta}}{\text{max }}\mathbb{E}_{c,x\sim X,X \sim \mathcal{D},t \sim \text{Uni}(\{1,2..T\}),i \sim \text{Uni}(\{1,2..N\}) }[\log \sigma(\gamma_X \beta'\log \frac{\pi_\theta(x_{t-1}|x_t,c)}{\pi_{\text{ref}}(x_{t-1}|x_t,c)}-\gamma_X \beta'\mu)]
        \label{eq:dpo_loss_pop_align_reduced}
\end{equation}

where $\text{Uni}()$ denotes the uniform distribution, $\gamma_X $ is an indicator with value +1 when $X$ is a winning population and -1 when $X$ is a losing population, $\beta' \propto \beta$ is a constant, $\mu$ is a normalizer, $x_t$ are sampled from a diffusion process. We provide a full proof of the derivation in Appendix~\ref{sec:proof}. This formulation allows us to train the model effectively without computing the whole diffusion process at each step. Empirically, we set $\mu=\mathbb{E}[\log \frac{\pi_\theta(x_{t-1}|x_t,c)}{\pi_{\text{ref}}(x_{t-1}|x_t,c)}]$ estimated through batch statistics.  

%% file: sections/4_results.tex
\section{Synthetic Evaluation}
To verify the behavior of our objective, we first conduct experiments on 1D mixture of Gaussians. In this simple setup, the reference distribution contains three Gaussians G1, G2, and G3, with a high skew between G1 and G3. G1, G3 is analogous to a pair of biased attributes such as "male", "female" where G3 is under-represented. G2 is analogous to an unrelated distribution, such as "trees" or "buildings". We collect 1000 samples to create a population-level preference dataset. The preference dataset do not contain samples from G2, just as our preference data do not contain non-human prompts.

We use \cref{eq:mixture_model} to represent $P_{\theta}$.
We initialize two models with $P_{\text{Ref}}$($w_{\text{Ref}}$ = \text{softmax}(1, 0, -1) , $\mu_{\text{Ref}}$ = (-7.0,0.0, 7.0), $\sigma_{\text{Ref}}$ = (1.0,1.0,1.0)) . We apply \ours (with $\beta$=0.5) and SFT loss to the models respectively and train the model until convergence. 

\begin{equation}
    P_{\theta} = \sum_{i=1}^{3} w_i \cdot \mathcal{N}(x; \mu_i, \sigma_i^2) \\  
    \text{ s.t.} \quad \sum_{i=1}^{3} w_i = 1, \quad \theta = \{w_i, \mu_i, \sigma_i^2 \mid i = 1, 2, 3\}
    \label{eq:mixture_model}
\end{equation}

\begin{figure}[t]
    \centering
    \begin{minipage}{0.49\linewidth}
        \centering
        \includegraphics[width=\linewidth]{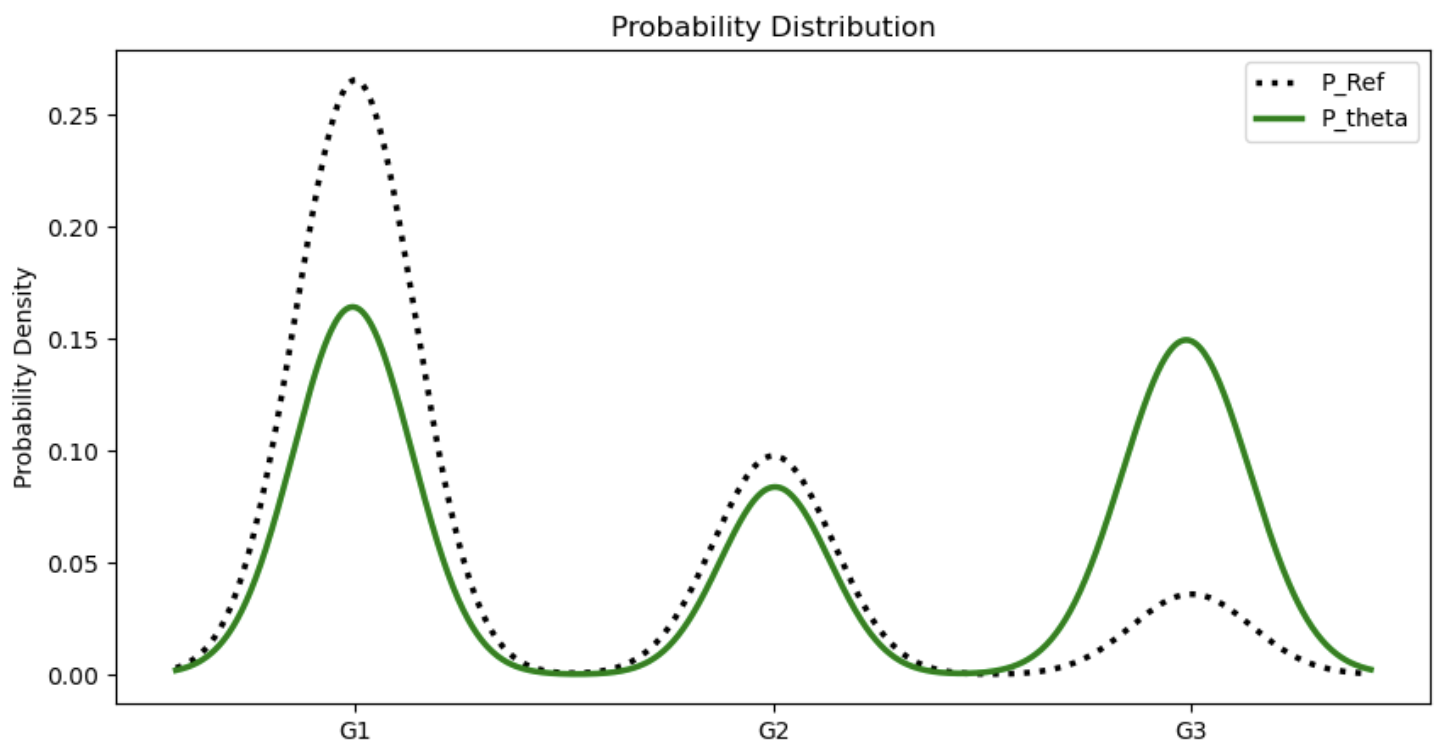}
        \caption*{(a) \ours }
    \end{minipage}
    \begin{minipage}{0.49\linewidth}
        \centering
        \includegraphics[width=\linewidth]{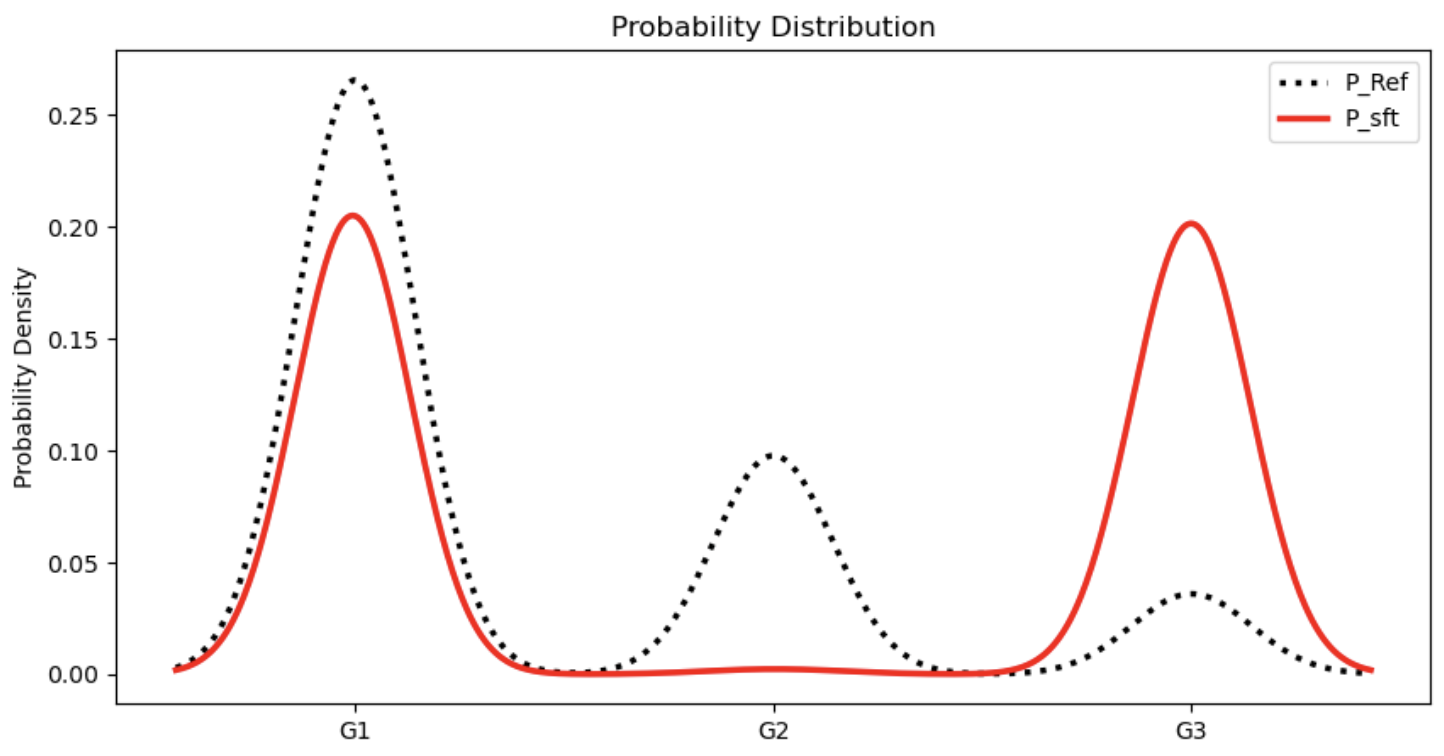}
        \caption*{(b) Supervised Fine Tuning (SFT)}
    \end{minipage}
    \caption{Effect of \ours and SFT on a skewed 1-d distribution \textbf{Left:} \ours effectively balance the skewed distribution. \textbf{Right:} Supervised Fine-Tuning (SFT) result in collapse of the Gaussian not in preference data }
    \label{fig:1d-exp}
\end{figure}

We show results in \cref{fig:1d-exp} we observe that \ours is able to mitigate the biases between G1,G3, while maintaining the distribution of G2. While SFT also balanced on G1,G3, it's support collapses on G2. These results are a simple illustration \ours's ability to mitigate the bias while maintaining the generative capability of the model gained from the pretraining data.

% skewed distribution of the generator as mixture of three gaussians G1, G2 and G3. G1 being the overrepresented class, G3 the underrepresented class and G2 being the other classes not involved in the skew.\\

% Firstly we construct Preference data using G1 and G3. We use \cref{eq:dpo_loss_pop_align_reduced} to get the skew balance between G1 and G3 as shown in \cref{fig:1d-exp} and still maintaining the distribution of G2, whereas using Supervised fine-tuning loss will result in the collapse of G2.\\

% We also observe the effect of $\beta$ and requirement of tuning $beta$ for optimal convergence. \cref{fig:beta-exp} shows that lower $\beta$ ($\beta$=0.1) results in overshooting of the under represented class, where higher $\beta$ ($\beta$=0.9) keeps it closer to the initial skew of $P_{ref}$.

\section{Experiments}

We conducted experiments with SDXL~\cite{podell2023sdxl}, a state-of-the-art T2I as the base model. We consider two aspects of biases: gender and race.  
% \ag{give 2-3 line intro.. we expt with SD version, sota model for image generation..align with respect to gender and race.. }
\subsection{Training Details}
\label{sec:training}
We use ChatGPT to generate 300 identity-neutral prompts involving no specific gender or race, such as "a botanist cataloging plant species in a dense forest" and "a biochemist examining cellular structures, in a high-tech lab". We augment the prompt with gender and race keywords as described in \cref{sec: data_pipeline}. In particular, we consider gender keywords "male" and "female" and race keywords "white","Asian","black","Latino Hispanic","Indian","middle eastern" as specified by the classifier. It should be noted that this list is not an exhaustive representation of all possible identities. However, our method can easily be generalized to incorporate other diversities with appropriate prompts. We generate 100 images for each identity-neutral prompts and 10 images for each identity-specific prompts. Afterwards, we obtain set-level preference data as described in \cref{sec: data_pipeline}. While images can be generated by either identity-neutral or identity-specific prompts in our pipeline, we use the identity-neutral prompt as the caption label in the training data.

We train our models using 4 Nvidia A5000 GPUs for 750 iterations. We use a per-GPU batch size of 2. We employ AdamW optimizer with a learning rate of 5e-07 for 750 iterations.

\subsection{Evaluation Metrics}

For fairness, we use the fairness discrepancy metric $f$ proposed by earlier works \cite{choi2020fair}, which measures fairness on sensitive attribute $u$ over individual image samples $x$ as 
\begin{equation}
    f (p_{\text{ref}} , p_{\theta} ) = \lVert \bbE_{p_{\text{ref}}}  [p(u|x)] -\bbE_{p_{\theta}}[p(u|x)] \rVert _2
    \label{eq:fairness}
\end{equation}
where $p_{\text{ref}}$ is an ideal distribution and $p_{\theta}$ is the distribution of a generative model. The lower is the discrepancy metric, the better can the model mitigate unfair biases. We use the DeepFace library, which contains various face detection and classification models for this metric \cite{serengil2024lightface1,serengil2020lightface2,serengil2021lightface3,serengil2023db4},

For image quality, we employ a set of standard image quality metrics: CLIP \cite{radford2021learning}, HPS v2 \cite{wu2023humanv2}, and LAION aesthetics score \cite{laion-aesthetics}.  CLIP measures the alignment of generated image and input prompts. LAION aesthetics score measures the quality of the generated image on its own. HPS takes into consider both the image quality and image-prompt alignment. For Pick-a-Pick benchmark, we additionally report PickScore \cite{kirstain2024pick}, which is trained on Pick-a-Pick dataset using human preference.  Higher values of these metrics indicates better quality of generated images.

% LIP \cite{radford2021learning} (MIT license), which measures image-text alignment, and PickScore \cite{kirstain2024pick}, HPS v2 \cite{wu2023humanv2} (Apache-2.0 license), and ImageReward \cite{xu2024imagereward} (Apache-2.0 license) which are caption-aware models that are trained to predict a human preference score given an image and its caption.

\begin{figure}[t]
    \centering
    \includegraphics[width=1.0\linewidth]{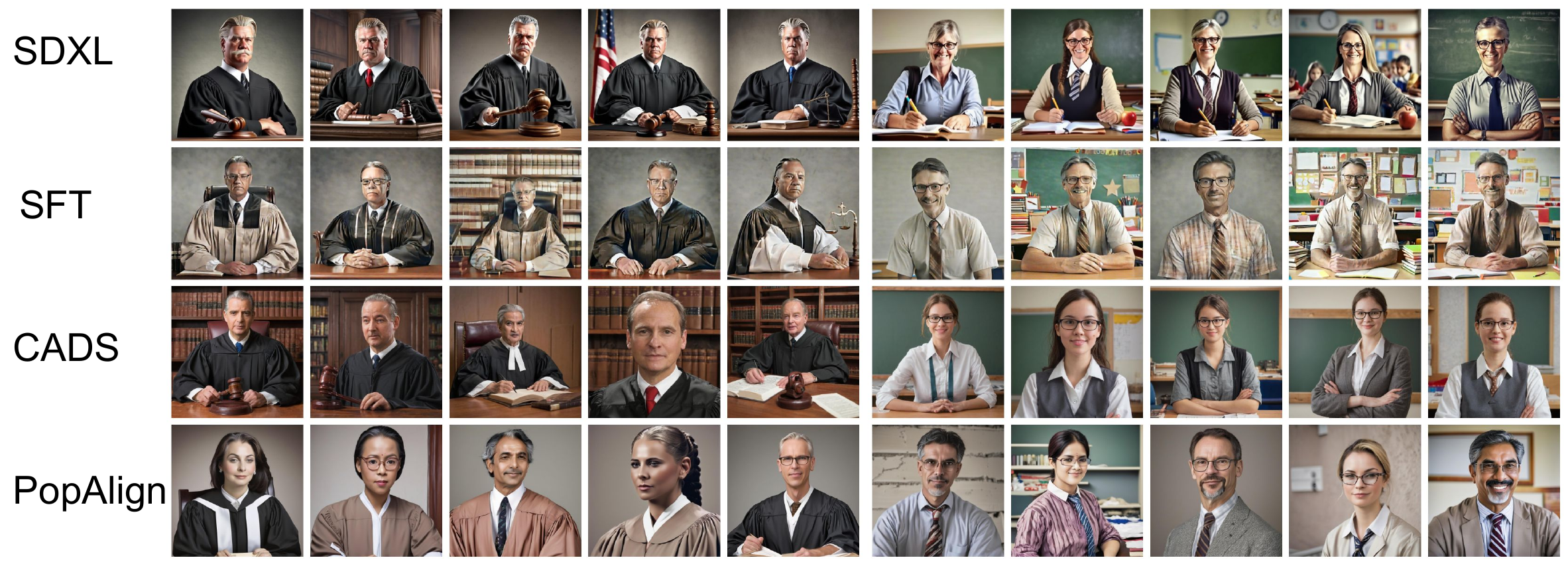}
    \caption{Qualitative results on gender-neutral prompts. \ours mitigates the bias of the pretrained SDXL in both male-skewed or female-skewed prompts.}
    \label{fig:qual_judge}
\end{figure}
\begin{table}[h!]
    \centering
    \caption{Results on gender-neutral and ethnic-neutral prompts.}
    \begin{tabular}{l cc ccc}
        \toprule
        & \multicolumn{2}{c}{Discrepancy} & \multicolumn{3}{c}{Quality} \\
        \cmidrule(r){2-3} \cmidrule(r){4-6}
        & Gender$\downarrow$ & Race$\downarrow$ & HPS $\uparrow$ & Aesthetic $\uparrow$ & CLIP $\uparrow$ \\
        \midrule
        SDXL & 0.417 & 0.666 & 25.2 ± 0.13 & 5.66 ± 0.01 & \textbf{28.2 ± 0.06} \\
        SDXL-SFT & 0.307 & 0.471 & 21.6 ± 0.11 & 5.72 ± 0.01 & 21.3 ± 0.05 \\
        SDXL-\ours & \textbf{0.184} & \textbf{0.258} & \textbf{25.9 ± 0.12} & \textbf{5.84 ± 0.01} & 28.2 ± 0.06 \\
        SDXL-CADS & 0.334 & 0.641 & 21.5 ± 0.15 & 5.83 ± 0.01 & 26.3 ± 0.05 \\
        SDXL-Dynamic-CFG & 0.307 & 0.552 & 22.5 ± 0.09 & 5.76 ± 0.01 & 26.4 ± 0.06 \\
        \hline
        SDXL-DPO & 0.294 & 0.642 & \textbf{34.6 ± 0.11} & 5.71 ± 0.01 & 31.5 ± 0.06 \\
        SDXL-DPO-\ours & 0.189 & \textbf{0.331} & 33.2 ± 0.12 & 5.84 ± 0.01 & 31.4 ± 0.04 \\
        \bottomrule
    \end{tabular}
    \label{tab:discrepancy}
\end{table}

\begin{table}[ht]
    \centering
    \caption{Results on identity-specific prompts.}
    \begin{tabular}{l ccc ccc}
        \toprule
        & \multicolumn{3}{c}{Recall} & \multicolumn{3}{c}{Quality} \\
        \cmidrule(r){2-4} \cmidrule(r){5-7}
        & Gender$\uparrow$ & Race$\uparrow$ & Overall$\uparrow$ & HPS$\uparrow$ & Aesthetic$\uparrow$ & CLIP$\uparrow$ \\
        \midrule
        SDXL & 100.0 & 99.8 & 99.8 & 36.7 ± 0.18 & 6.05 ± 0.01 & 33.6 ± 0.11 \\
        SDXL-SFT & 100.0 & 95.1 & 95.1 & 35.6 ± 0.17 & 5.96 ± 0.01 & 33.1 ± 0.12 \\
        SDXL-PopAlign & 99.0 & 98.8 & 98.0 & 36.8 ± 0.18 & 6.09 ± 0.01 & 33.4 ± 0.11 \\
        \midrule
        SDXL-DPO & 99.8 & 99.8 & 99.6 & 38.2 ± 0.17 & 6.20 ± 0.01 & 33.8 ± 0.12 \\
        SDXL-DPO-PopAlign & 100.0 & 99.8 & 98.8 & 37.8 ± 0.18 & 6.27 ± 0.01 & 33.5 ± 0.11 \\
        \bottomrule
    \end{tabular}
    \label{tab:recall}
\end{table}

\begin{table}[ht]
    \centering
    \caption{Results on generic prompts from Pick-a-Pick test set. These prompts are not necessarily gender-neutral and ethnic-neutral.\ours was able to maintain the image quality on generic prompts.}
   
\begin{tabular}{l c ccc}
        \toprule
        Model & PickScore$\uparrow$ & HPS $\uparrow$ & Aesthetic $\uparrow$ & CLIP  $\uparrow$ \\
        \midrule
        SDXL & 21.9 ± 0.06 & 36.2 ± 0.23 & 5.87 ± 0.02 & 32.8 ± 0.15 \\
        SDXL-SFT & 21.3 ± 0.05 & 33.9 ± 0.24 & 5.76 ± 0.02 & 31.6 ± 0.16 \\
        SDXL-PopAlign & 21.9 ± 0.05 & 35.4 ± 0.20 & 5.89 ± 0.02 & 32.3 ± 0.15 \\
        \midrule
        SDXL-DPO & 22.3 ± 0.05 & 37.2 ± 0.22 & 5.89 ± 0.02 & 33.4 ± 0.15 \\
        SDXL-DPO-PopAlign & 22.4 ± 0.03 & 37.2 ± 0.19 & 5.90 ± 0.01 & 33.2 ± 0.11 \\
        \bottomrule
    \end{tabular}
    
    \label{tab:pick}
\end{table}

\begin{figure}[t]
    \centering
    \includegraphics[width=0.8\linewidth]{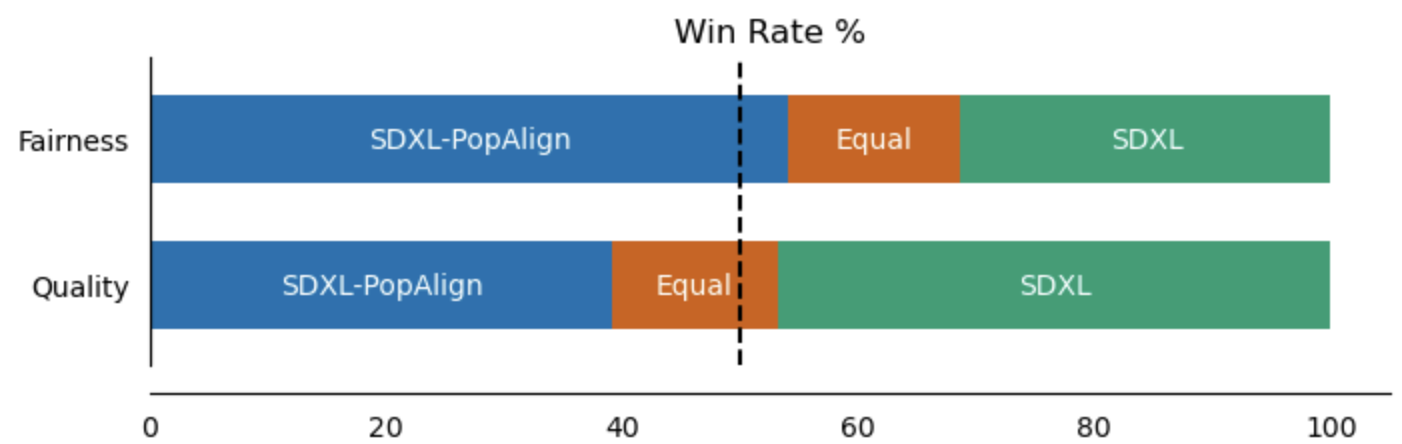}
    \caption{Human Evaluation on fairness and quality of the image population}
    \label{fig:human_eval}
\end{figure}

\subsection{Identity-Neutral Prompts}
\label{sec: identity_neutral}
We evaluate the performance of our method on a set of 100 gender neural prompts. These prompts are manually written and are different from the training prompts. To minimize detection and classification errors, we use simple prompts with the template "best quality, a realistic photo of [identity-neutral prompt]". We use simple prompts that do not involve multiple persons to reduce potential errors in classification results. For each prompt, we generate 100 images, achieving a total sample size of 10,000. We report the discrepancy metric on gender and race, as well as image quality metrics HPS v2, LAION aesthetic score, and CLIP.  We train \ours using both the standard SDXL as the starting point, as well as a SDXL released by Diffusion-DPO \cite{wallace2023diffusion}. We compare against supervised fine-tuning (SDXL-SFT) baseline, which naively fine-tune the diffusion model on the "winning" sets of generated images.  We also compare against CADS and Dynamic-CFG \cite{sadat2023cads}, which are training-free methods to improve sample diversity. 

These results are shown in \cref{tab:discrepancy}. Amongst all compared methods,  SDXL-PopAlign achieves the lowest discrepancy metric. Notably, SDXL-PopAlign reduces the gender and race discrepancy of the pretrained SDXL by (-0.233), and (-0.408) respectively, while maintaining comparable image quality as measured by HPS, Aesthetic, and CLIP scores.  Similarly, when initializing with a DPO checkpoint, PopAlign was able to reduce the gender and race discrepancy by (-0.105) and (-0.311) respectively, while maintaining comparable image quality. Thanks to alignment on human preference, SDXL-DPO has a higher image quality than SDXL as measured by HPS, Aesthetic, and CLIP scores. SDXL-DPO-PopAlign is able to maintain such a lead while reducing the biases of the model significantly.

We also provide qualitative results in \cref{fig:qual_judge}. In this example, we show images generated of SDXL-PopAlign, SDXL, and SDXL-SFT using the same prompt "judge". Visually,  PopAlign generate a more diverse set of identity than SDXL, and SDXL-SFT. In terms of quality,  SDXL-SFT was able to improve the perceived fairness of SDXL at the cost of generation quality. SDXL-PopAlign was able to maintain the image quality of SDXL, while offering a more diverse set of generations in terms of gender and ethnicity. 

As classifiers are not perfect, we also conducted human evaluations. We ask the user to judge the fairness of quality of images generated by SDXL and SDXL-PopAlign. The images are grouped into sets of 5 images. We show the results in \cref{fig:human_eval}. Humans generally consider PopAlign a superior model in terms of fairness, and the two models are roughly comparable in terms of image quality.

\subsection{Identity-Specific Prompts}
\label{sec: identity_specific}
To verify that our model do not over-generalize superficial diversity for identity-specific prompts, we evaluate our method against the pretrained model and SFT baseline on  a set of identity-specific prompts. This is crucial because a model that misrepresents a particular identity when explicitly prompted to do so will raise equity and fairness concerns and is not safe to deploy in an end-user product. We create this specific prompts by augmenting the identity neutral prompts in \cref{sec: identity_neutral} with identity keywords such as "female", "Asian". To measure the image-prompt alignment, we report the recall rate of gender and race classifier. Specifically, we classify each of the generated images and check if the classification results match the prompt. We also report image quality metrics including HPS v2, LAION aesthetics and CLIP. We show these results in \cref{tab:recall}. 

Almost all methods achieve high scores in recall metrics, suggesting training to mitigate biases on identity-neutral prompts do not adversely affect the generation results of identity-specific prompts. However, SDXL-SFT suffers a slightly larger drop in the overall recall than SDXL-PopAlign. In terms of image quality, we observed a similar pattern as in identity-specific prompts, where PopAlign better preserve than image quality of pretrained models than SFT baselines, as measured in HPS (+1.5), Aesthetic (+0.13) and CLIP (+0.7). 

\subsection{Generic Prompts in the Wild}

For optimal classification accuracy, we use simple prompts for experiments in \cref{sec: identity_neutral} and \cref{sec: identity_specific}. However, these prompts are vastly different than complicated prompts typically used by human users. To provide a more comprehensive evaluation of generation quality of PopAlign and pretrained SDXL, we evaluate our models on Pick-a-Pick test set \cite{kirstain2024pick} consisting of prompts written by actual humans. These prompts are not necessarily identity-neutral. In fact, some prompts do not include humans at all. In addition to HPS, LAION aesthetics and CLIP metrics, we additionally report the PickScore which is commonly used on this benchmark.  We show results in \cref{tab:pick}. These results are consistent with previous experiments. SDXL-PopAlign was able to match the performance of pretrained SDXL, and achieves higher image quality of than SFT baselines.

%% file: sections/5_conclusion.tex
\section{Limitations }
\label{sec:limitations}

Our method can only mitigate the biases to a certain degree. It cannot completely eliminate all perceived biases. In general, there is a trade-off between fairness and image quality, as shown in our extensive ablations. The user can adjust these parameters based on how much they value these two goals with respect to each other. 

Additionally, our method assumes all prompts that do not explicitly includes gender or identity as neutral prompts. However, people may have varying views. For example, people may disagree on if "the president of the United States" should leads to images of a female president. On one hand, one should not assume the leader of a free democratic society be limited to a specific gender. On the other hand, at the time of writing there is no female president of the United States. In this aspect, generating an image of female president may be considered as a misrepresentation of fact, which can hardly be called "fair". This is especially the case for prompts involving a historic context, like "the president of the United States in the 1800s". We avoid using these potentially controversial prompts.

Our model relies on gender and race classifier which achieves high performance over the categories on which they are trained. However, there are ethnicity in the real world beyond the fixed set of classes. Similarly, our gender classifier fails to represent the existence of non-binary gender. We have proposed a pipeline to collect preference data for bias mitigation using human feedback. In principle, it should be able to curate a preference dataset representing these nuances with human annotators. Due to the prohibitively expensive cost, we left these for future works to address.

Additionally, there is also the concern that if the visual appearance should dictate a person's gender and ethnicity as opposed to self-identification. In this aspect, our model can only identify "gender appearances" and "ethnic appearances", but not "gender identities" and "ethnic identities" as these concepts involves non-visual elements such as self-recognition.

\section{Broader Impacts}
\label{sec:border_impacts}
\ours~ aims to reduce certain commonly perceived biases on text-to-image generative models, such as gender and racial biases. \ours~ can be particular to useful as an extra step before the  release of new T2I models to mitigate the biases without sacrificing image quality. However, it may also inadvertently perpetuate new biases, such as non-binary genders and minority races, which could be excluded from the preference datasets. Therefore, we suggest users to take extra caution when dealing with these situations. 

As any other image generator, PopAlign may be misused to create realistic-looking images for deception, fraud and other illegal activities. In addition, by adjusting the preference data, an adversary may use PopAlign to amplify existing gender and ethnical biases, such as creating an image model generating exclusively light-skinned characters. We do not condone these kinds of use.

\section{Conclusions}

In summary, we propose \ours, a novel algorithm that mitigates the biases of pretrained text-to-image diffusion models while preserving the quality of the generated images. \ours successfully extend the pair-wise preference formulation used by RLHF and DPO to a novel population-level alignment objective, surpassing comparable baselines in both human evaluation and quantitative metrics. In particular, \ours~ outperforms the supervised fine-tuning baseline on identity-neutral prompts, identity-specific prompts, as well as generic human written prompts in terms of both fairness and image quality. However, it is also important to recognize that our experiments are limited in that it employs a race-gender classifier that assumes a binary gender categorization and a limited set of races. It does not capture the complicated nuances such as non-binary gender identity and many under-represented races.  We plan to address these limitations in future works by employing real humans to create a more diverse set of training data that capture these nuances. 
\label{sec:conclusion}

% \ag{need a broader implications section for this paper. check if it counts towards page limit. if yes, move to appendix. write at least two paras on this. see some other fair generation papers for good examples. make sure to discuss both pros and cons. eg, bad actors could collect even more biased preference datasets to further harm the fairness of generative models. }

%% file: sections/6_appendix.tex
\appendix
% \setcounter{section}{0}
% \renewcommand{\thesection}{A.\arabic{section}}
% \setcounter{equation}{0}
% \renewcommand{\theequation}{A\arabic{equation}}
% \setcounter{table}{0}
% \renewcommand{\thetable}{A\arabic{table}}
% \setcounter{figure}{0}
% \renewcommand{\thefigure}{A\arabic{figure}}

% % \section*{Appendix}
% \large Appendix
% \label{sec:appendix}

\section{Ablation Studies}
\label{sec:ablation}
To validate our design choices, we conducted extensive ablation studies on various hyper-parameters.

\subsection{Classifier-Free Guidance }

Classifier free guidance (CFG) is the used to ensure the generated images accurately follow the text prompts. Typically, higher guidance strength leads to sharper images and better image-prompt alignment, at the cost of sample diversity. We show effects of varying CFG on identity-neutral prompts in \cref{fig:abltaion_cfg}. For SDXL, higher CFG leads to higher discrepancy, indicating less diversity as expected. However, for SFT and PopAlign, increasing CFG do not significantly compromise the discrepancy because of extra training. Among these two methods, PopAlign consistently exhibits a lower discrepancy. For main experiments, we used a cfg of 6.5.

\subsection{Divergence Penalty }

\begin{figure}
\centering
\begin{minipage}{.5\textwidth}
  \centering
  \includegraphics[width=1.0\linewidth]{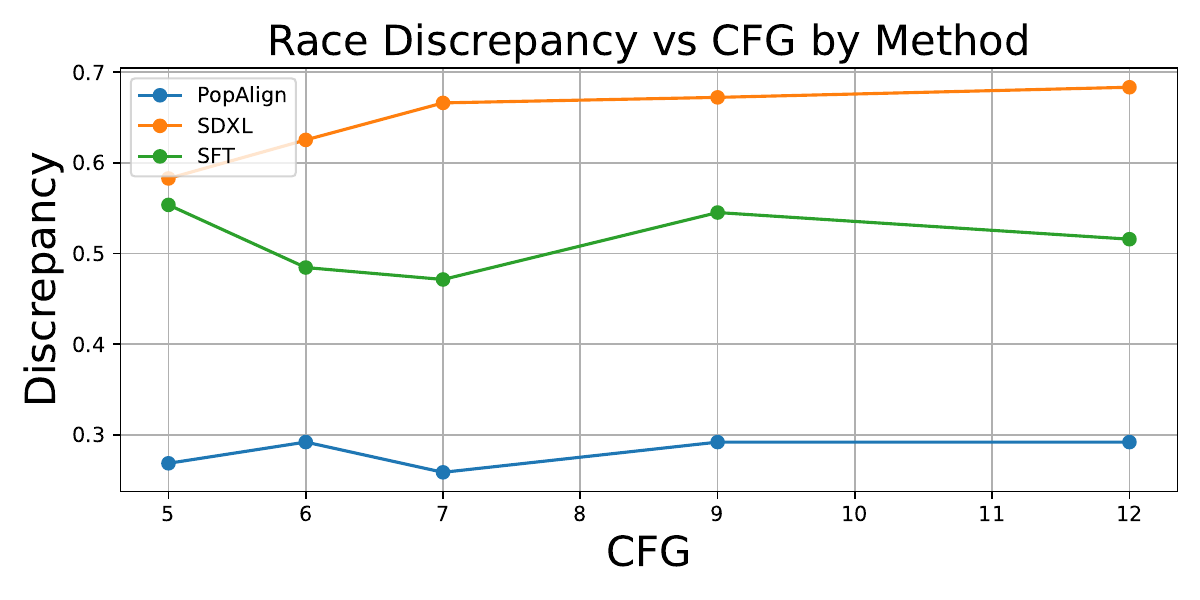}
  \captionof{figure}{Ablation study of varying CFGs}
  \label{fig:abltaion_cfg}
\end{minipage}%
\begin{minipage}{.5\textwidth}
  \centering
  \includegraphics[width=1.0\linewidth]{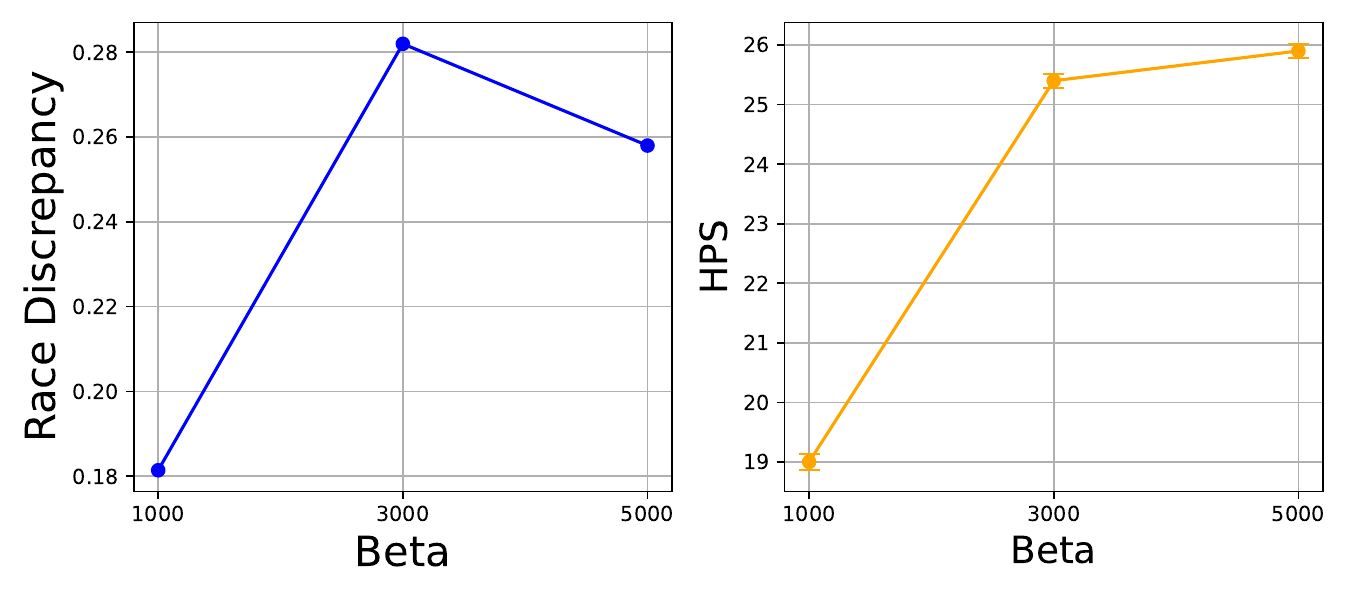}
  \captionof{figure}{Ablation study on divergence penalty $\beta$}
  \label{fig:ablation_beta}
\end{minipage}
\end{figure}

The divergence Penalty $\beta$ is an important hyperparameter as it controls the strength of divergence penalty.  We show the results of $\beta=1000$, $\beta=3000$ and $\beta=5000$ in \cref{fig:ablation_beta}. In general, higher $\beta$ leads to higher image quality as stronger divergence penalty prevents the model from deviating too much from the pretrained checkpoint. This comes with a cost of higher discrepancy. We pick $\beta=5000$ for our experiments, but end-users may choose an alternative based on the relative importance of fairness and image quality.   

\subsection{Normalization Factor}
% \subsubsection{AI Feedback}
% \subsubsection{Human Feedback}
Because we remove pair wise preferences in \cref{eq:dpo_loss_pop_align_reduced}, we need to center the inner term (reward) by $\mu$. Following the conventional practice of RL, we use the expected value of inner term $\mu=\mathbb{E}[\log \frac{\pi_\theta(x_{t-1}|x_t,c)}{\pi_{\text{ref}}(x_{t-1}|x_t,c)}]$, which can be re-written as the weight sum of the expected reward of all positive samples and that of all negative samples $\mu(\alpha)=\alpha\mathbb{E}[\log \frac{\pi_\theta(x_{t-1}^w|x_t^w,c)}{\pi_{\text{ref}}(x_{t-1}^w|x_t^w,c)}]+(1-\alpha)\mathbb{E}[\log \frac{\pi_\theta(x^l_{t-1}|x^l_t,c)}{\pi_{\text{ref}}(x^l_{t-1}|x^l_t,c)}]$ with $\alpha=0.5$.   We also experimented with two alternatives $\alpha=0.25$ and $\alpha=0.75$. $\alpha=0.25$ will move the $\mu$ closer to the side of losing samples, while  $\alpha=0.75$ will move the $\mu$  closer to the side of winning samples. Since the gradient of $\log \sigma$ is symmetric with respect to the origin, and monotonically decreases as it moves away from the origin. $\alpha=0.25$ will increase the update step of the negative samples because $\mu(0.25)$ is closer to the negative samples, which makes the inner term closer to the origin. Similarly, $\mu(0.25)$ will increase the update step of the positive samples. We show the results in \cref{tab:reference}. 

$\alpha=0.25$ leads to model divergence, as the negative samples have a stronger "pushing force" than the "pulling force" of positive samples in this setup. $\alpha=0.75$ leads to lower discrepancy and image quality, as it increases the "pulling force" of positive samples, implicitly decreasing the effect of divergence penalty. 

\begin{table}[ht!]
    \centering
    \caption{Ablation study of normalization factor.}
    \begin{tabular}{l cc ccc}
        \toprule
        & \multicolumn{2}{c}{Discrepancy} & \multicolumn{3}{c}{Quality} \\
        \cmidrule(r){2-3} \cmidrule(r){4-6}
        $\alpha$ & Gender$\downarrow$ & Race$\downarrow$ & HPS $\uparrow$ & Aesthetic $\uparrow$ & CLIP $\uparrow$ \\
        \midrule
        0.25  & Diverge & Diverge & -0.5 & 4.25 & 16.3 \\
        0.5  & 0.184 & 0.258 & \textbf{25.9} & \textbf{5.84} & \textbf{28.2} \\ 
        0.75 & \textbf{0.170} & \textbf{0.222} & 23.7 & 5.72 & 26.4 \\ 
        \bottomrule
    \end{tabular}
    
    \label{tab:reference}
\end{table}

\section{Human-in-the-Loop Sampling}
\label{sec:human_in_the_loop}
Our method make uses of a pretrained classifier. However, this representation over simplifies the complex nuances of the real world. In this section, we discuss a human-in-the loop sampling process that may be used to capture these complex aspects of diversity (e.g. non-binary genders, homosexual versus heterosexual couples). 

In particular, we consider the setup where human start with a set of initial samples $X^0=\{x^0_1...x^0_N\}$ where N is the set size. A Markov chain iteratively refines the set of samples $X^t$ to $X^{t+1}$ until it reaches a target distribution.  At each step, a random sample $x^t_i$ is replaced with a new sample $x^{t+1}_i$, and we obtain a new set $X^{t+1}=\{x^t_1...x^t_{i-1},x^{t+1}_i,x^t_{i+1},x^t_N\}$. We accept the new set if the new set is closer to the target distribution, and back-track otherwise. In the human-in-the-loop version, a human judge will accept the changes that improve the perceived "fairness". For example, in a male-dominate population, the judge will accept changes that replace a male sample with a female sample, and reject changes in the other direction. The judge will terminate the process when we achieved a fair balance between the male and female samples. Assuming humans and classifiers are reasonably capably of distinguishing the basic properties in question, this human-in-the-loop MCMC should generate a fair distribution with sufficiently large samples as the training data. However, because finding human annotators qualified to fairly evaluate these nuances and curate a large collection of data can be prohibitively expansive, we left this for future works.

% \section{Implementation Details of 1D-Gaussian Experiments}

% We use \cref{eq:mixture_model} to represent $P_{\theta}$.
% We initialize two models with $P_{\text{Ref}}$( $w_{\text{Ref}}$ = \text{softmax}(1, 0, -1) , $\mu_{\text{Ref}}$ = (-7.0,0.0, 7.0), $\sigma_{\text{Ref}}$ = (1.0,1.0,1.0). We apply PopAlign(with $\beta$=0.5) and SFT loss to the models respectively and train the model until convergence. 

% \begin{equation}
%     P_{\theta} = \sum_{i=1}^{3} w_i \cdot \mathcal{N}(x; \mu_i, \sigma_i^2) \\  
%     \text{ s.t.} \quad \sum_{i=1}^{3} w_i = 1, \quad \theta = \{w_i, \mu_i, \sigma_i^2 \mid i = 1, 2, 3\}
%     \label{eq:mixture_model}
% \end{equation}

\section{Proof of Population Level Alignment Objective}
\label{sec:proof}
We start with \cref{eq:dpo_loss_pop}. Following Diffusion-DPO \cite{wallace2023diffusion}, we can substitute  $\pi_\theta(x|c)$ with $\sum_{t=1}^T\pi_\theta(x_t|x_{t+1},c)$ and obtain

\begin{align}
 &=\mathbb{E}_{c,X^w,X^l \sim \mathcal{D}}[\log \sigma(\beta \log \frac{\prod_{i=1}^n\prod_{t=1}^T\pi_\theta(x^{w,i}_t|x^{w,i}_{t+1},c)}{\prod_{i=1}^n\prod_{t=1}^T\pi_{\text{ref}}(x^{w,i}_t|x^{w,i}_{t+1},c)}-\\ & \qquad\qquad\qquad\qquad\qquad\qquad\qquad\qquad\qquad\qquad\beta \log \frac{\prod_{i=1}^n\prod_{t=1}^T\pi_\theta(x^{l,i}_t|x^{l,i}_{t+1},c)}{\prod_{i=1}^n\prod_{t=1}^T\pi_{\text{ref}}(x^{l,i}_t|x^{l,i}_{t+1},c)})] \\ 
 &= \mathbb{E}_{c,X^w,X^l \sim \mathcal{D}}[\log \sigma(\beta  \sum_{i=1}^N\sum_{t=1}^T\log\pi_\theta(x^{w,i}_t|x^{w,i}_{t+1},c) -\beta\sum_{i=1}^N\sum_{t=1}^T\log\pi_{\text{ref}}(x^{w,i}_t|x^{w,i}_{t+1},c)-\\  &\qquad \qquad\qquad\qquad\qquad\beta  \sum_{i=1}^N\sum_{t=1}^T\log\pi_\theta(x^{l,i}_t|x^{l,i}_{t+1},c) +\beta{\sum_{i=1}^N\sum_{t=1}^T\log\pi_{\text{ref}}(x^{l,i}_t|x^{l,i}_{t+1},c)})]
        \label{eq:dpo_loss_pop_deri2}
\end{align}

By Jensen's inequality, we have a lower bound 

\begin{align}
     &\mathbb{E}_{c,X^w,X^l \sim \mathcal{D}}[\log \sigma(\beta  \sum_{i=1}^N\sum_{t=1}^T\log\pi_\theta(x^{w,i}_t|x^{w,i}_{t+1},c) -\beta\sum_{i=1}^N\sum_{t=1}^T\log\pi_{\text{ref}}(x^{w,i}_t|x^{w,i}_{t+1},c)-\\  &\qquad \qquad\qquad\beta  \sum_{i=1}^N\sum_{t=1}^T\log\pi_\theta(x^{l,i}_t|x^{l,i}_{t+1},c) +\beta{\sum_{i=1}^N\sum_{t=1}^T\log\pi_{\text{ref}}(x^{l,i}_t|x^{l,i}_{t+1},c)})] \\
     & \geq  n T \mathbb{E}_{c,x^w,x^l \sim \mathcal{D},t\in \text{Unif}(\{1,2..T\}),i\in \text{Unif}(\{1,2..N\})}[\log \sigma(\beta  \log\pi_\theta(x^{w,i}_t|x^{w,i}_{t+1},c) -\\  &\qquad \qquad\qquad\beta  \log\pi_{\text{ref}}(x^{w,i}_t|x^{w,i}_{t+1},c)-\beta\log\pi_\theta(x^{l,i}_t|x^{l,i}_{t+1},c) +\beta\log\pi_{\text{ref}}(x^{l,i}_t|x^{l,i}_{t+1},c))]\\
     &=  n T \mathbb{E}_{c,x^w,x^l \sim \mathcal{D},t\in \text{Unif}(\{1,2..T\}),i\in \text{Unif}(\{1,2..N\})}[\log \sigma(\beta  \log\frac{\pi_\theta(x^{w,i}_t|x^{w,i}_{t+1},c)}{\pi_{\text{ref}}(x^{w,i}_t|x^{w,i}_{t+1},c)}-\\& \qquad\qquad  \qquad\beta\log\frac{\pi_\theta(x^{l,i}_t|x^{l,i}_{t+1},c)} {\pi_{\text{ref}}(x^{l,i}_t|x^{l,i}_{t+1},c)})]\\
          &=  n T \mathbb{E}_{c,x^w,x^l \sim \mathcal{D},t\in \text{Unif}(\{1,2..T\}),i\in \text{Unif}(\{1,2..N\})}[\log \sigma(\beta  \log\frac{\pi_\theta(x^{w,i}_t|x^{w,i}_{t+1},c)}{\pi_{\text{ref}}(x^{w,i}_t|x^{w,i}_{t+1},c)}-\\& \qquad\qquad  \qquad \mu + \mu - \beta\log\frac{\pi_\theta(x^{l,i}_t|x^{l,i}_{t+1},c)} {\pi_{\text{ref}}(x^{l,i}_t|x^{l,i}_{t+1},c)})]\\
                   &\leq 2 N T \mathbb{E}_{c,x  \sim X, X  \sim \mathcal{D},t\in \text{Unif}(\{1,2..T\}),i\in \text{Unif}(\{1,2..N\})}[\log \sigma(\\& \qquad\qquad  \qquad \gamma_X \beta  \log\frac{\pi_\theta(x_t|x_{t+1},c)}{\pi_{\text{ref}}(x_t|x_{t+1},c)} - \gamma_X \beta \mu)]
                   \\
                   &= \mathbb{E}_{c,x  \sim X, X  \sim \mathcal{D},t\in \text{Unif}(\{1,2..T\}),i\in \text{Unif}(\{1,2..N\})}[\log \sigma( \gamma_X \beta'  \log\frac{\pi_\theta(x_t|x_{t+1},c)}{\pi_{\text{ref}}(x_t|x_{t+1},c)} - \gamma_X \beta' \mu )]
        \label{eq:dpo_loss_pop_deri3}
\end{align}

where $\beta'=2NT\beta$ and $\mu$ is a normalizing constant to stabilize the optimization.

\section{Details of Human Evaluation}
\label{sec:human_eval}
We use the following prompt for human evaluation
\begin{tcolorbox}[colframe=black, title=Human Evaluation Prompt]
Select the set of images that represents more diversity of identity representation and quality of image.\\

Look at the two sets of images below generated from a prompt. Each set contains multiple images. Set A is the top 5 images, while Set B is the bottom 5. Select which set you think shows greater diversity in terms of identity representation and quality of image set.\\

Please consider the variety in elements such as color, subject matter, race, gender, and other visible identity markers when making your selection.\\

\textbf{Which set is more diverse and fair in terms of identity representation:}\\

- Set A (Top Row) is more diverse and fair

- Set B (Bottom Row) is more diverse and fair

- Both sets are equally diverse and fair\\

\textbf{Which set has better quality images overall:}\\

- Set A (Top Row) is higher quality

- Set B (Bottom Row) is higher quality

- Both sets are equally good in terms of quality
\end{tcolorbox}

For each pair of sets, we collect responses from three individual human evaluators to mitigate potential noises in human preference. We do not expose human evaluators for any NSFW content. We employ Amazon MTurk for this job. The works are paid with a prorated hourly minimum wage. We follow all guidelines and rules of respective institutions.

\section{Additional Qualitative Results}

\begin{figure}[t]
    \centering
    \includegraphics[width=\linewidth]{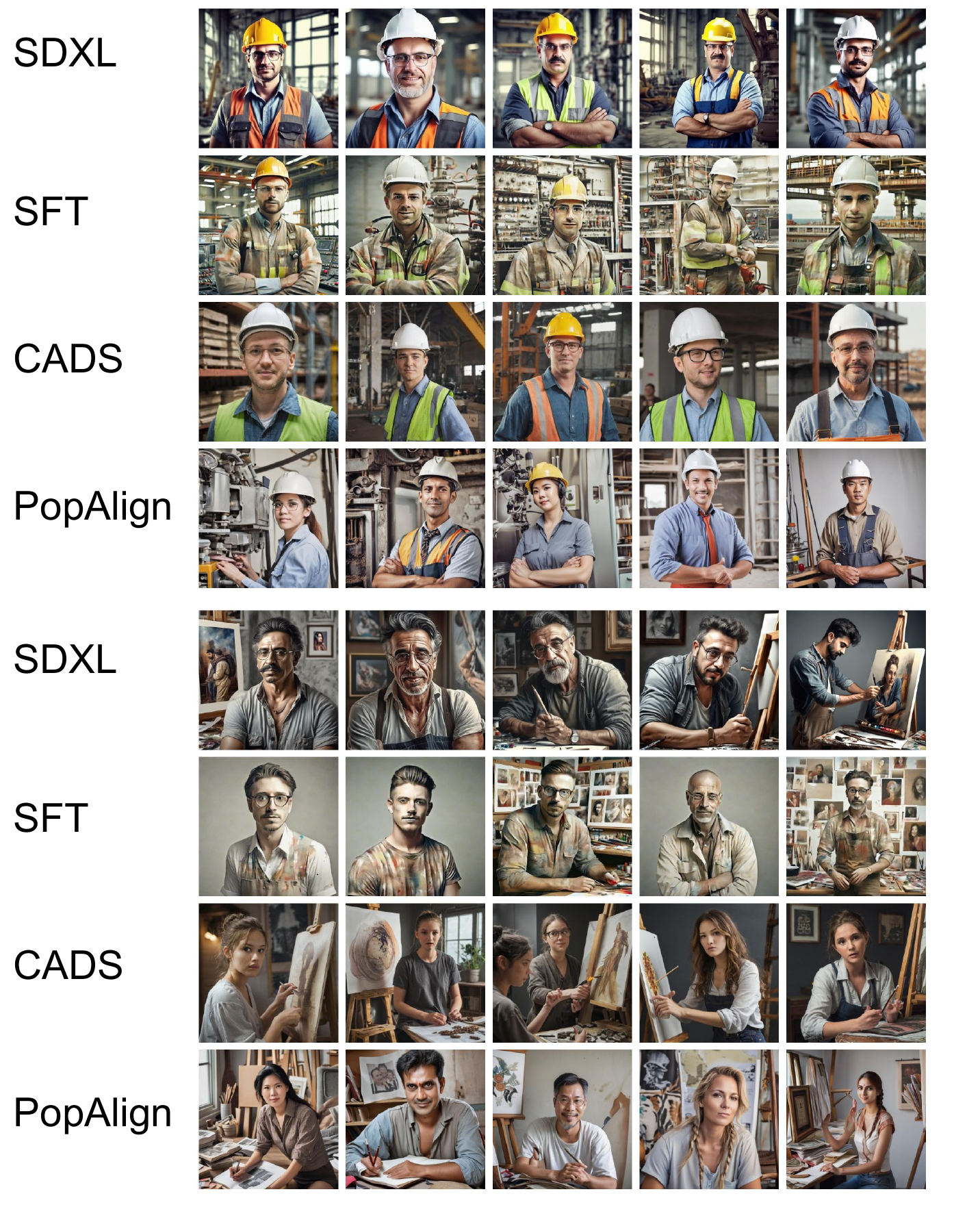}
    \caption{\textbf{Additional qualitative results on gender-neutral prompts.} \ours~ offers a diverse representation of identities while maintaining a comparable image quality with the SDXL baseline. The top four rows are generated using the prompt "engineer", while the bottom four rows are generated using the prompt "artist". The prompts are formatted in "best quality, a realistic photo of [prompt]" }
    \label{fig:additional_vis}
\end{figure}

We provide additional qualitative results in \cref{fig:additional_vis}. The samples are generated using the prompt "engineer" and "artist". Compared with the baselines,  \ours~ offers a diverse representation of identities while maintaining a comparable image quality with the pretrained SDXL checkpoint.

% \section{Additional Qualitative Results}
% \begin{figure}[h]
%     \centering
%     \includegraphics[width=\linewidth]{figures/appendix-sidebyside2.pdf}
%     \caption{Caption}
%     \label{fig:enter-label}
% \end{figure}

\section{Additional Experiments}

\begin{figure}[t]
    \centering
    \begin{subfigure}[t]{0.32\linewidth}
        \centering
        \includegraphics[width=\linewidth]{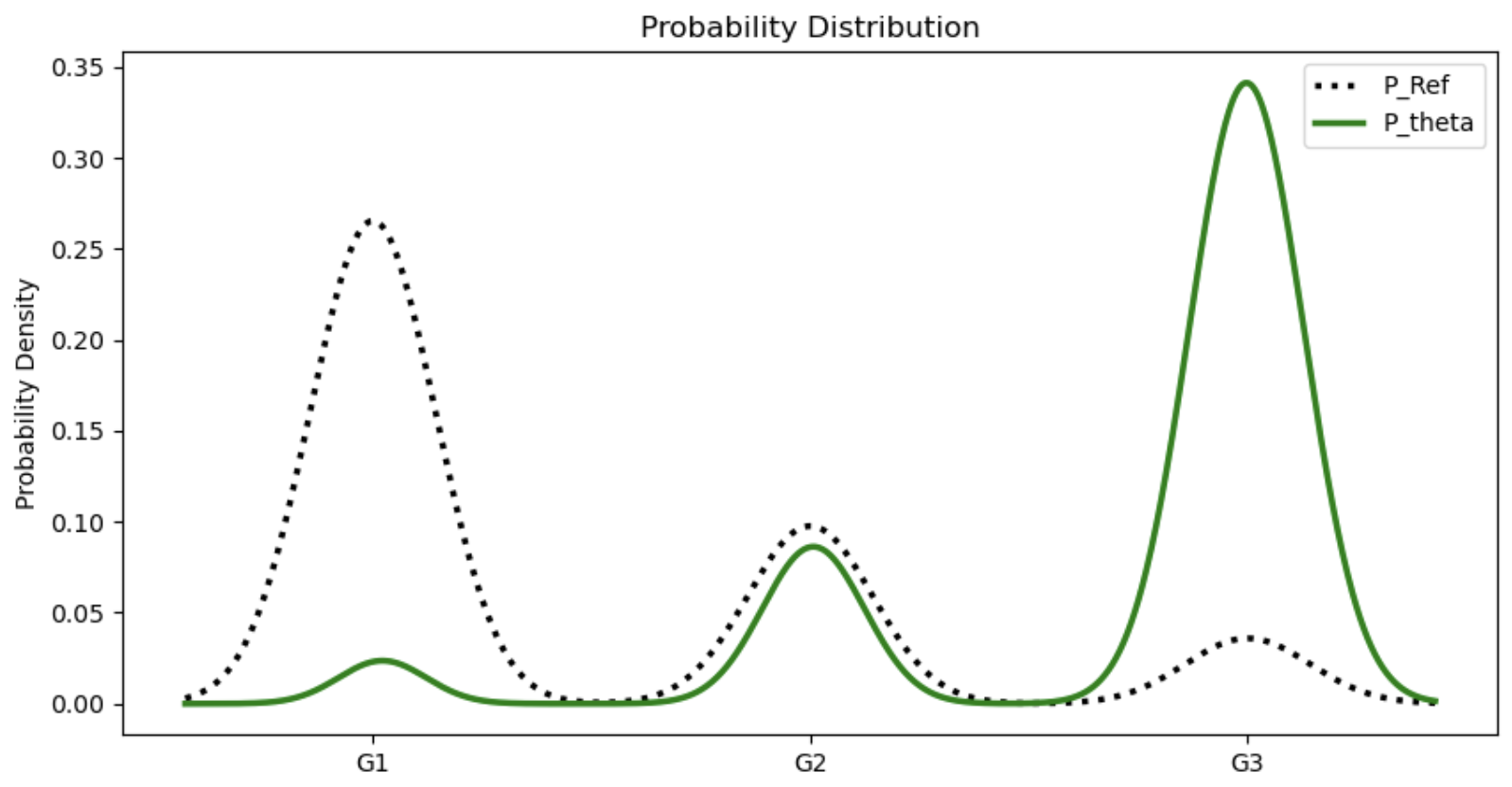}
        \caption{\ours~ with $\beta$=0.1}
    \end{subfigure}
    \begin{subfigure}[t]{0.32\linewidth}
        \centering
        \includegraphics[width=\linewidth]{figures/DPO_0.5.png}
        \caption{\ours~ with $\beta$=0.5}
    \end{subfigure}
    \begin{subfigure}[t]{0.32\linewidth}
        \centering
        \includegraphics[width=\linewidth]{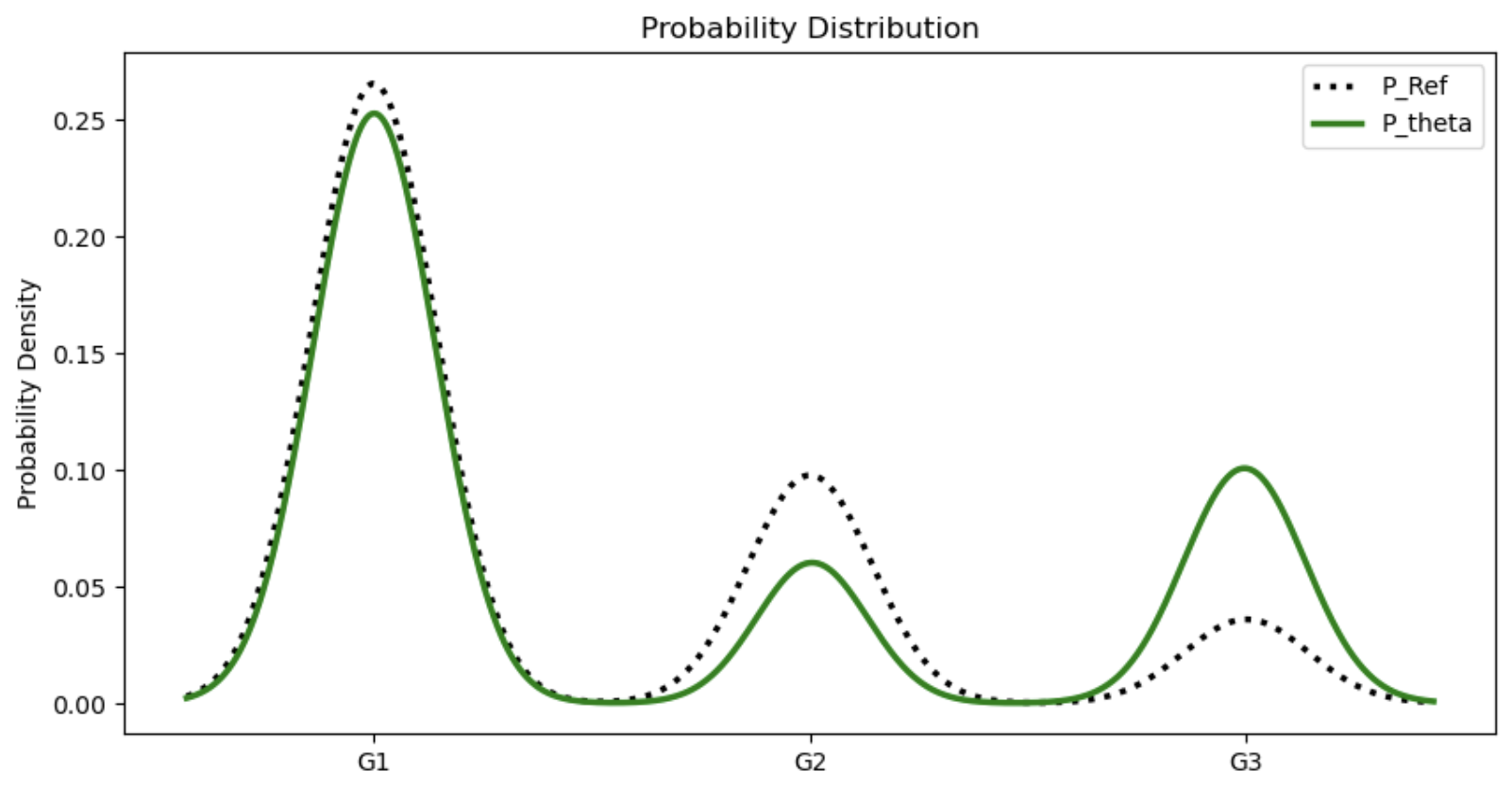}
        \caption{\ours~ with $\beta$=0.9}
    \end{subfigure}
    \caption{Effect of $\beta$ in \ours~}
    \label{fig:beta-exp}
\end{figure}

On 1D mixture of Gaussian, the divergence penalty $\beta$ is an important hyperparameter as it controls the strength of divergence penalty.  We show the results of $\beta=0.1$, $\beta=0.5$ and $\beta=0.9$ in \cref{fig:beta-exp}. In general, higher $\beta$ leads to stronger divergence penalty prevents the model from deviating too much from the pretrained checkpoint. This comes with a cost of higher discrepancy. We pick $\beta=0.5$ for our 1-d experiments.   

\begin{figure}[t]
    \centering
    \begin{subfigure}[t]{0.32\linewidth}
        \centering
        \includegraphics[width=\linewidth]{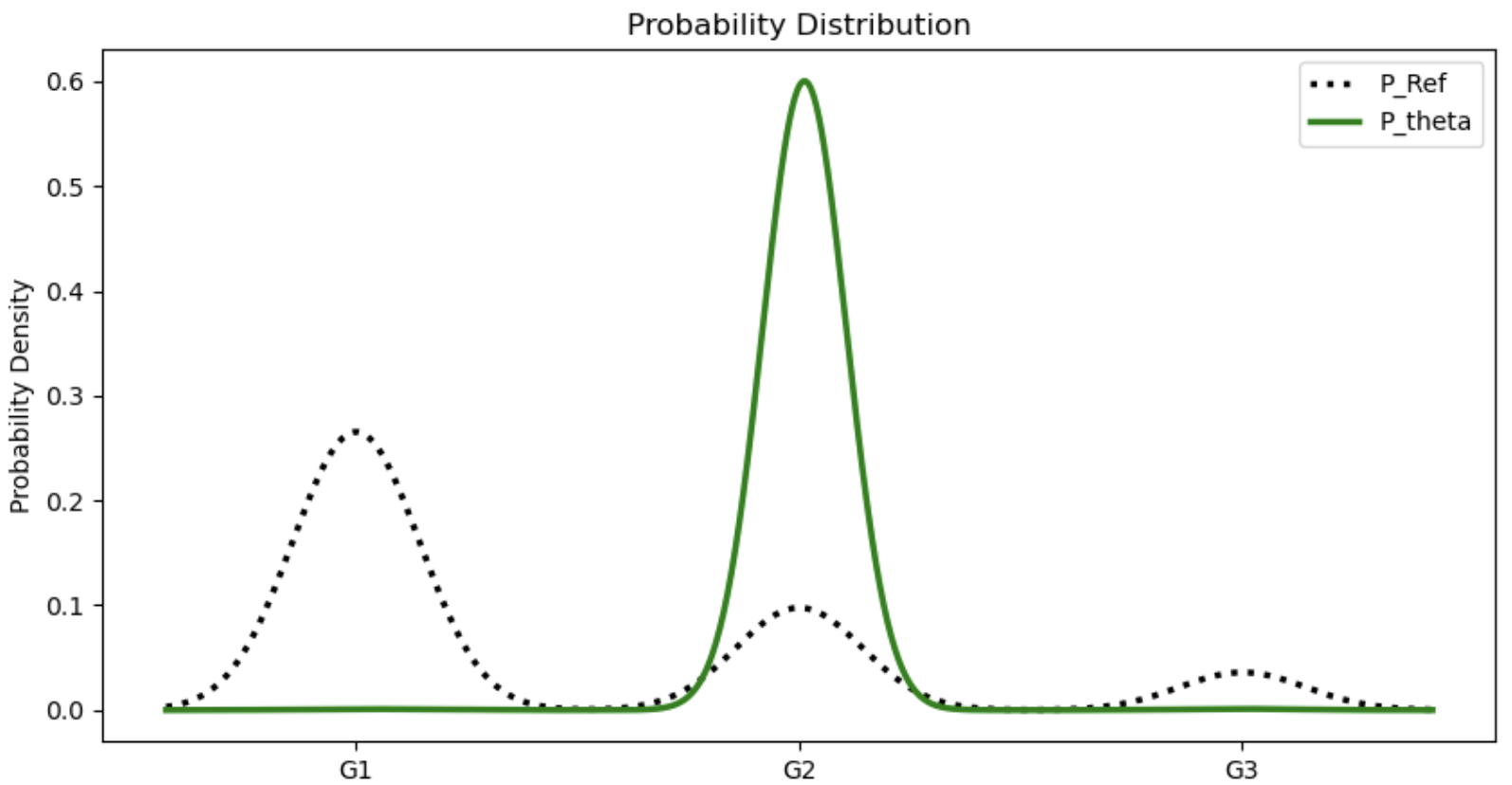}
        \caption{\ours~ with $\alpha$=0.25}
    \end{subfigure}
    \begin{subfigure}[t]{0.32\linewidth}
        \centering
        \includegraphics[width=\linewidth]{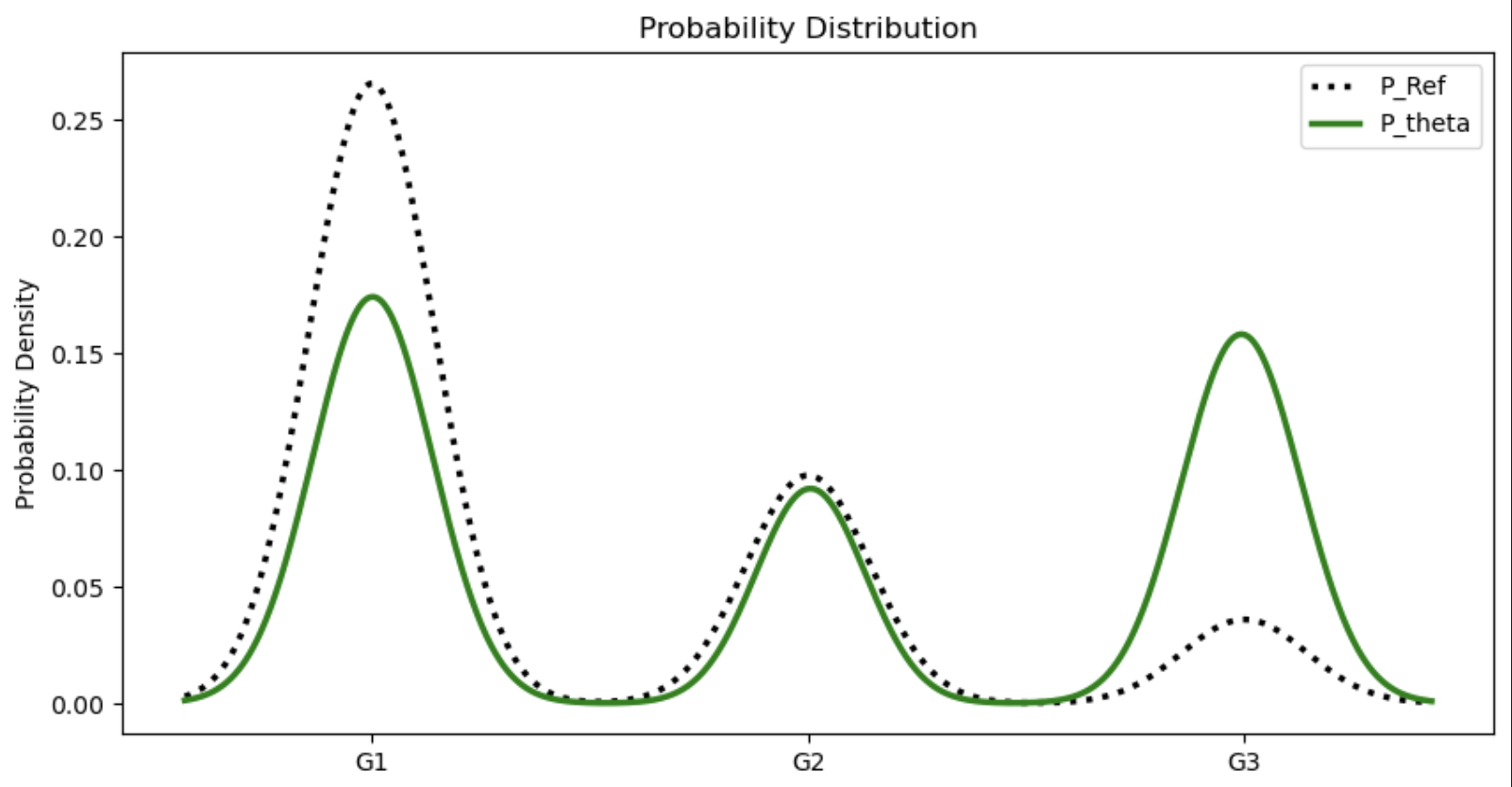}
        \caption{\ours~ with $\alpha$=0.5}
    \end{subfigure}
    \begin{subfigure}[t]{0.32\linewidth}
        \centering
        \includegraphics[width=\linewidth]{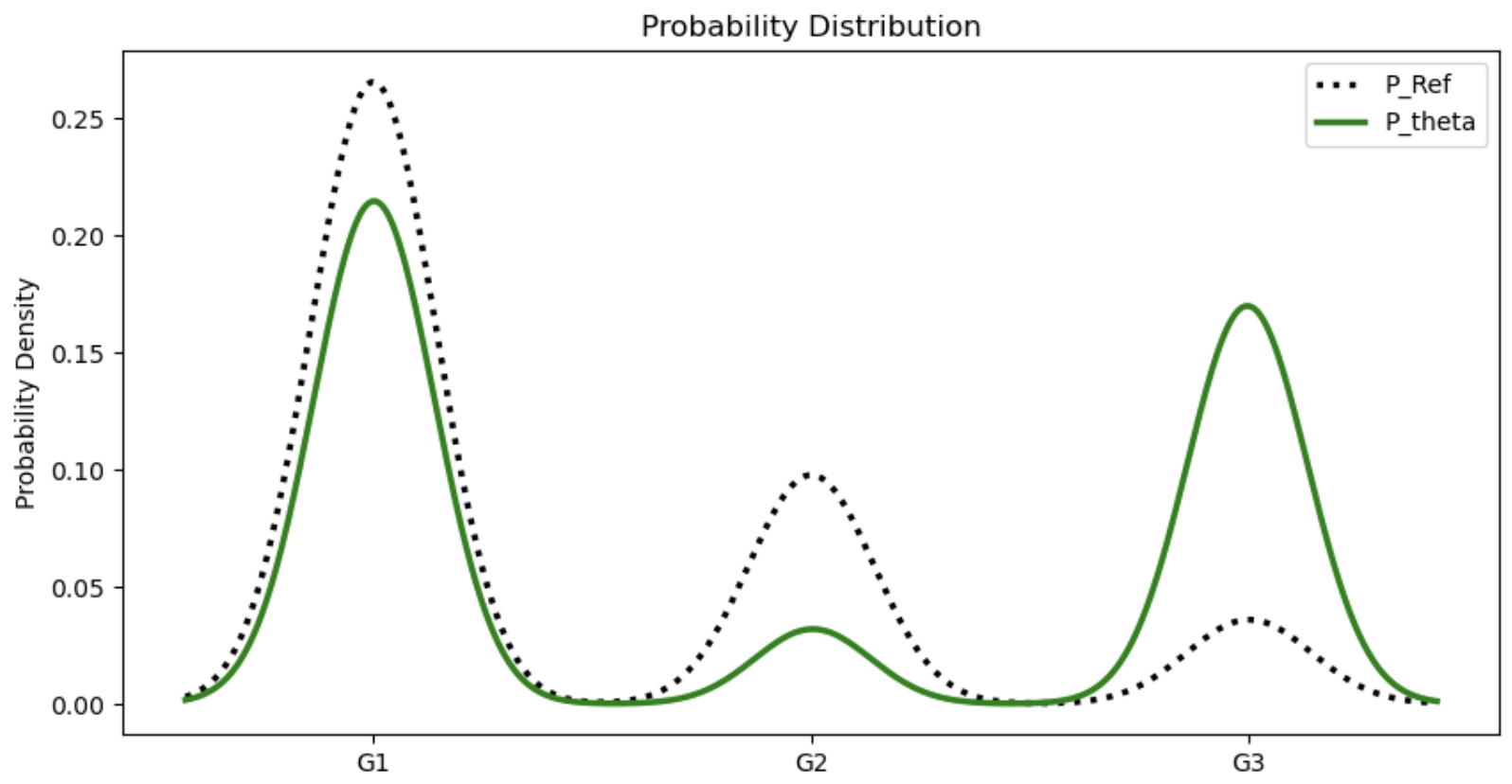}
        \caption{\ours~ with $\alpha$=0.75}
    \end{subfigure}
    \caption{Effect of $\alpha$ in \ours~}
    \label{fig:alpha-exp}
\end{figure}

 We also experimented with $\alpha=0.25$, $\alpha=0.5$, $\alpha=0.75$, $\alpha=0.25$ will move the $\mu$ closer to the side of losing samples, while  $\alpha=0.75$ will move the $\mu$  closer to the side of winning samples. Since the gradient of $\log \sigma$ is symmetric with respect to the origin, and monotonically decreases as it moves away from the origin. $\alpha=0.25$ will increase the update step of the negative samples because $\mu(0.25)$ is closer to the negative samples, which makes the inner term closer to the origin. Similarly, $\mu(0.25)$ will increase the update step of the positive samples. We show the results in \cref{fig:alpha-exp}. 

$\alpha=0.25$ leads to model divergence, as the negative samples have a stronger "pushing force" than the "pulling force" of positive samples in this setup. $\alpha=0.75$ leads to lower discrepancy, as it increases the "pulling force" of positive samples, implicitly decreasing the effect of divergence penalty. 

% \section{Training to mitigate gender biases only.}
\section{Statistic Significance}
\label{sec:error_bars}
We report the 95\% confidence interval for the main experiments reported in \cref{tab:discrepancy,tab:pick,tab:recall}. The confidence interval is computed using the standard error of the sample mean estimator using s-statistics.

\section{Licenses}
\label{sec:licenses}
We makes use the following models: CLIP (MIT license), PickScore(MIT license), HPS v2 (Apache-2.0 license), LAION Aesthetics predictor (MIT license), Deepface (MIT license), SDXL(CreativeML Open RAIL++-M License). Diffusion-DPO (Apache-2.0 license). 

We use prompts from Pick-a-Pick dataset (MIT License).

\section{Safe Guards}
\label{sec:safeguard}
\ours is based on the diffuser \cite{diffusers2023} library. It should be used with the standard safeguards such as NSFW safety checker and hidden watermarks. For the released prompt, we manually inspected them and found no harmful content.

\section{Human Subjects}
Human evaluation conducted in this study has received exemption from UCLA IRB.